\documentclass[english]{IEEEtran}
\usepackage[T1]{fontenc}
\usepackage[latin9]{inputenc}
\usepackage{float}
\usepackage{amsthm}
\usepackage{amsmath}
\usepackage{amssymb}
\usepackage{graphicx}

\makeatletter

\newcommand{\noun}[1]{\textsc{#1}}
\providecommand{\tabularnewline}{\\}

 \usepackage[ruled,vlined]{algorithm2e}
\theoremstyle{plain}
\newtheorem{thm}{\protect\theoremname}
\theoremstyle{definition}
\newtheorem{problem}[thm]{\protect\problemname}
\theoremstyle{plain}
\newtheorem{cor}[thm]{\protect\corollaryname}
\theoremstyle{plain}
\newtheorem{prop}[thm]{\protect\propositionname}
\newenvironment{lyxcode}
{\par\begin{list}{}{
\setlength{\rightmargin}{\leftmargin}
\setlength{\listparindent}{0pt}
\raggedright
\setlength{\itemsep}{0pt}
\setlength{\parsep}{0pt}
\normalfont\ttfamily}%
 \item[]}
{\end{list}}

\usepackage{babel}
\usepackage{times}
\usepackage{xcolor}
\usepackage[T1]{fontenc}

\DeclareMathOperator{\interior}{int}

\newcommand{\dobst}{\Delta}

\newcommand{\trajectorystore}{trajectory store}

\newcommand{\ts}{H}

\@ifundefined{showcaptionsetup}{}{%
 \PassOptionsToPackage{caption=false}{subfig}}
\usepackage{subfig}
\makeatother

\usepackage{babel}
\providecommand{\corollaryname}{Corollary}
\providecommand{\problemname}{Problem}
\providecommand{\propositionname}{Proposition}
\providecommand{\theoremname}{Theorem}

\begin{document}

\title{Prioritized Planning Algorithms for Trajectory Coordination of Multiple
Mobile Robots}

\author{Michal {\v C}{\' a}p, Peter Nov{\' a}k, Alexander Kleiner, Martin Seleck{\' y}}
\maketitle
\begin{abstract}
An important capability of autonomous multi-robot systems is to prevent
collision among the individual robots. One approach to this problem
is to plan conflict-free trajectories and let each of the robots follow
its pre-planned trajectory. A widely used practical method for multi-robot
trajectory planning is prioritized planning, which has been shown
to be effective in practice, but is in general incomplete. Formal
analysis of instances that are provably solvable by prioritized planning
is still missing. Moreover, prioritized planning is a centralized
algorithm, which may be in many situations undesirable. 

In this paper we a) propose a revised version of prioritized planning
and characterize the class of instances that are provably solvable
by the algorithm and b) propose an asynchronous decentralized variant
of prioritized planning, which maintains the desirable properties
of the centralized version and in the same time exploits the distributed
computational power of the individual robots, which in most situations
allows to find the joint trajectories faster.

The experimental evaluation performed on real-world indoor maps shows
that a) the revised version of prioritized planning reliably solves
a wide class of instances on which both classical prioritized planning
and popular reactive technique ORCA fail and b) the asynchronous decentralized
algorithm provides solution faster than the previously proposed synchronized
decentralized algorithm.\\

\emph{Note to Practitioners}---Consider a large warehouse in which
the goods are stored and retrieved by autonomous transport robots.
One option is to ignore interaction between the vehicles during the
route planning for each robot and handle the conflicts only during
the route execution. However, such approach is prone to deadlocks,
i.e. to a situations during which some of the robots mutually block
each other, cannot proceed and fail to complete their transportation
task. An alternative approach would involve planning collision-free
routes for each robot before the robots start executing them. However,
the general methods for this problem that guarantee a solution are
impractical due to their high computational complexity. In this paper,
we show that a simple prioritized approach in which robots plan their
routes one after another is guaranteed to find collision-free trajectories
for a well-defined class of practical problems. In particular, if
the systems resembles human-made transport infrastructures by requiring
that the start and destination position of each vehicle must never
obstruct other vehicles from moving, then the proposed approach is
guaranteed to provide a solution. E.g., in such a warehouse application,
the collision-free routes can efficiently computed by the prioritized
approach. The paper contains formal condition that characterizes the
problem instances for which is the method guaranteed to work.

Further, we propose a new decentralized adaptation of the prioritized
algorithm that can be used in multi-robot systems without a central
solver. This technique can be used to find coordinate trajectories
just by running simple negotiation protocol between the individual
robots. The paper contains analysis showing that the decentralized
algorithm exhibits desirable theoretical properties and experimental
comparison of performance of different variations of centralized and
decentralized algorithms.\end{abstract}

\section{Introduction}

When mobile robots autonomously operate in a shared space, one of
the crucial issues is how to coordinate the trajectories of individual
robots so as to prevent potential collisions. The two most commonly
used classes of methods that deal with this problems are multi-robot
planning and reactive collision avoidance.

Multi-robot motion planners take into consideration the start and
goal position of each robot and plan coordinated trajectories that
are mutually conflict-free. If the robots execute the resulting joint
plan precisely (or within some given tolerance), it is guaranteed
that the robots will reach their goals without collision. However,
it is known that even the simplest variants of multi-robot path planning
problem are intractable. Deciding whether a coordinated collision-free
paths exist for multiple discs moving amidst polygonal obstacles is
known to be strongly NP-hard~\cite{SpirakisY84_Strong_NP_Hardness_of_Moving_Many_Discs};
the same task involving rectangular objects in an empty room is known
to be in PSPACE-hard~\cite{hopcroft84}. 

The multi-robot planners are typically based either on the coupled
heuristic search in the joint state space of all robots or on decoupled
planning. The coupled approaches typically find optimal solutions
\cite{Standley10,StandleyK11,WagnerC11_Mstar}, but do not scale well
with the increasing number of conflicting robots. 

On the other hand, decoupled approaches plan independently for each
robot. They can be fast enough for real-time applications, but they
typically suffer from incompleteness. 

A widely used decoupled scheme for the multi-robot motion planning
that has been shown to be effective in practice is prioritized planning~\cite{Erdmann87onmultiple}.
In prioritized planning, each robot is assigned a unique priority
and the algorithm proceeds sequentially from the highest priority
robot to the lowest priority one. At each iteration, one of the robots
plans its trajectory such that it avoids the higher-priority robots.
Such a greedy approach is clearly incomplete if we allow arbitrary
maps and arbitrary start and goal locations for each robot, but in
relatively sparse environments, the techniques tends to perform well. 

Recently, Velagapudi et al. presented a decentralized version of prioritized
planning technique for teams of mobile robots~\cite{VelagapudiSS10},
which is able to utilize the distributed computational resources to
reduce the time needed to find a solution. Since the algorithm proceeds
in globally-synchronized rounds, faster-computing robots have to wait
at the end of each round for the longest-computing robot and thus
the distributed computational power may not be used efficiently.

The contribution of this paper is twofold: 1)~We propose a revised
version of the prioritized planning scheme and show that for this
revised version it is possible to provide sufficient conditions under
which is the algorithm guaranteed to provide a solution. 2)~We propose
an novel asynchronous decentralized variant of both classical and
revised prioritized planning scheme that is guaranteed to terminate
and inherits completeness properties from the respective centralized
counterpart. We experimentally show that asynchronous decentralized
algorithm exhibits better utilization of the distributed computational
resources and thus provides faster convergence times compared to the
previously presented synchronized approach. Finally, we demonstrate
a practical applicability of the asynchronous approach in a cooperative
multi-UAV scenario.

Partial results of the presented work appeared in \cite{cap_2012_adpp,cap_2013_b},
where the focus was on the design of asynchronous version of decentralized
prioritized planning. Here, we extend our previous work by proposing
the revised version of prioritized planning scheme, by theoretical
analysis of the properties of all discussed algorithms, by performing
experimental comparison on real-world indoor maps and by including
reactive techniques into the comparison.

\section{\label{sec:Problem-Definition}Problem Definition}

Consider $n$ circular robots operating in a 2-d workspace $\mathcal{W}\subseteq\mathbb{R}^{2}$.
The subset of $\mathcal{W}$ occupied by the body of robot $i$ when
its center is on position $x$ is denoted as $R_{i}(x)$. The maximum
speed the robot $i$ can move at is denoted as $v_{i}$. Each robot
is assumed to be assigned a \emph{task} that involves moving from
its start position $s_{i}$ to some goal position $g_{i}$ and stay
there. We assume that the start and goal positions of all robots are
mutually disjunct, i.e. the bodies of robots do not overlap when the
robots are on their start positions and when they are on their goal
positions.

A path $p:\:[0,1]\rightarrow\mathcal{W}$ of robot $i$ in workspace
$\mathcal{W}$ is called \emph{satisfying }if it starts at the robot's
start position $s_{i}$, ends at robot's goal position $g_{i}$, and
the body of robot whose center follows the path $p$ always lies entirely
in $\mathcal{W}$. A trajectory $\pi:\:[0,\infty)\rightarrow\mathcal{W}$
is a mapping from time points to positions in workspace and unlike
a path it carries information about how it should be executed in time.
Analogically, a trajectory of robot $i$ is called \emph{satisfying
}if it starts at the robot's start position $s_{i}$, finally reaches
and stays at the goal position $g_{i}$, the body of robot $i$ whose
center follows the trajectory $\pi$ always lies entirely in $\mathcal{W}$,
and the robot never moves faster than its maximum speed $v_{i}$.

The trajectories $\pi_{i},\pi_{j}$ of two robots $i,j$ are said
to be \emph{conflict-free} if the bodies of the robots $i,j$ never
intersect when they follow the trajectories $\pi_{i}$ and $\pi_{j}$. 
\begin{problem}
[Trajectory Coordination Problem] \label{problem:TrajectoryCoordinationProblem}Given
a workspace $\mathcal{W}$ and tasks $\left\langle s_{1},g_{1}\right\rangle ,\ldots,\left\langle s_{n},g_{n}\right\rangle $
for robots $1,\ldots,n$, find trajectories $\pi_{1},\ldots,\pi_{n}$
such that each trajectory $\pi_{i}$ is satisfying for robot $i$
and trajectories $\pi_{i},\pi_{j}$ of every two different robots
$i,j$ are mutually conflict-free.
\end{problem}

\subsection*{Notation}

The following shorthand notations will be used to talk about regions
occupied by a different subsets of robots at their start and goal
positions: 

\[
\begin{array}{rclcrcl}
S^{i} & := & R_{i}(s_{i}) &  & G^{i} & := & R_{i}(g_{i})\\
S^{>i} & := & \underset{j=i+1,\ldots,n}{\cup}R_{j}(s_{j}) &  & G^{<i} & := & \underset{j=1,\ldots,i-1}{\cup}R_{j}(g_{j})\\
S & := & \underset{j=1,\ldots,n}{\cup}R_{j}(s_{j}) &  & G & := & \underset{j=1,\ldots,n}{\cup}R_{j}(g{}_{j})
\end{array}
\]

Further, we will work with the concept of a space-time region: When
a spatial object, such as the body of a robot, follows a given trajectory,
then it can be thought of as occupying a certain region in space-time
$\mathcal{T}:=\mathcal{W}\times\left[0,\infty\right)$. A dynamic
obstacle $\Delta$ is then a region in such a space-time $\mathcal{T}$.
If $(x,y,t)\in\Delta$, then we know that the spatial position $(x,y)$
is occupied by dynamic obstacle $\Delta$ at time $t$. The function
\[
R_{i}^{\Delta}(\pi):=\left\{ (x,y,t):t\in[0,\infty)\wedge(x,y)\in R_{i}(\pi(t))\right\} 
\]
 maps trajectories of a robot $i$ to regions of space-time that the
robot $i$ occupies when its center point follows given trajectory
$\pi$. As a special case, let $R_{i}^{\Delta}(\emptyset):=\emptyset$.

\subsection*{Assumptions on Communication}

We assume that each robot is equipped with an independent computation
unit and a wireless device for communication with other robots. Wireless
communication channels are typically implemented as broadcast channels,
where each communicated message is broadcast, but ignored by the nodes
that are not among the declared recipients of the message. In such
a channel a single broadcast message uses the same channel capacity
as a single point-to-point message and thus we will prefer to perform
a single broadcast instead of sending several point-to-point messages.
Further, in the following discussion we will assume that such a communication
channel is reliable, i.e. each broadcast messages is eventually received
by all robots in the system, and that the communication channel preserves
the ordering of messages that were sent in.

\section{\label{sec:Prioritized-Planning-1}Prioritized Planning}

A straightforward approach to solve the trajectory coordination problem
would be to see all robots in the system as one composite robot with
many degrees of freedom and use some path planning algorithm to find
a joint path for all the robots. However, the size of such a joint
configuration space is exponential in the number of robots and thus
this approach quickly becomes impractical if one wants to plan for
more than a few robots. A pragmatic approach that is often useful
even for large multi-robot teams is prioritized planning. The idea
has been first articulated by Erdman and Lozano-P�rez in \cite{Erdmann87onmultiple}.
Other works such as \cite{BergO05,Bennewitz02Planning} investigate
techniques for choosing a good prioritization for the robots.

\subsection*{Classical Prioritized Planning}

In prioritized planning each robot is assigned a unique priority.
The trajectories for individual robots are then planned sequentially
from the highest priority robot to the lowest priority one. For each
robot a trajectory is planned that avoids both the static obstacles
in the environment and the higher-priority robots moving along the
trajectories planned in the previous iterations. The pseudocode of
classical prioritized planning is in Algorithm~\ref{alg:PP}.

\SetKwProg{alg}{Algorithm}{}{}
\SetKwProg{procedure}{Procedure}{}{}
\SetKwProg{function}{Function}{}{}
\SetKwFunction{besttraj}{Best-traj}
\SetKwFunction{pp}{PP}
\LinesNumbered
\begin{algorithm}
\alg{\pp}{

	$\dobst\leftarrow\emptyset$\;

	\For{$i\gets1\ldots n$}{

		$\pi_{i}\gets$\besttraj{$\mathcal{W},\dobst$}\;\label{alg:cpp-best-traj-1}

		\If{$\pi_{i}=\emptyset$}{

		 report failure and terminate

		}

		$\dobst\leftarrow\dobst\cup R_{i}^{\Delta}(\pi_{i})$\;

		}

}

\function{\besttraj{$\mathcal{W}',\dobst$}}{

	return optimal satisfying trajectory for robot $i$ in $\mathcal{W}'$
that avoids regions $\dobst$ if it exists, otherwise return $\emptyset$

}

\caption{\label{alg:PP} Classical Prioritized Planning}
\end{algorithm}

The algorithm iterates over the robots, starting from the highest-priority
robot $1$ to the lowest-priority robot $n$. During $i$-th iteration
the algorithm computes a trajectory for robot $i$ that avoids the
space-time regions occupied by robots $1,\ldots,i-1$.  

The trajectory of robot $i$ is computed in \besttraj{$\mathcal{W}',\dobst$}
function. The function returns a trajectory for robot $i$ such that
body of the robot always stays inside the static workspace $\mathcal{W}'$
and avoids dynamic regions $\dobst$ occupied by other robots. Such
a function would be in practice implemented using some application-specific
technique for motion planning with dynamic obstacles, e.g. \cite{vanBerg2007kinodynamic_planning_with_dynamic_obst}
or \cite{narayanan2012anytime_interval_planning}. As it will become
clear later, it is desirable that this function is implemented using
an algorithm that offers some form of completeness, since this property
will be inherited also by the multi-robot algorithm.

\subsubsection*{Properties}

The algorithm terminates either with success or with failure in at
most $n$ iterations. The successful termination occurs in exactly
$n$ iterations if valid trajectories for all robots have been found.
The termination with failure occurs if there exists a robot for whom
no satisfying trajectory that avoids higher-priority robots have been
found.

When the algorithm terminates successfully, each robot is assigned
a trajectory that is conflict-free with the trajectories of all other
robots. This follows from the fact that the final trajectory of each
robot $i$ is conflict-free with the higher-priority robots, since
robot $i$ avoided collision with them and also with lower-priority
robots since they avoided conflict with the trajectory of robot $i$
themselves. 

Prioritized planning is in general incomplete, consider the counter-example~\cite{Silver05}
depicted in Figure~\ref{fig:Instance_where_PP_fails}:

\begin{center}
\begin{figure}[H]
\begin{centering}
\includegraphics[scale=0.5]{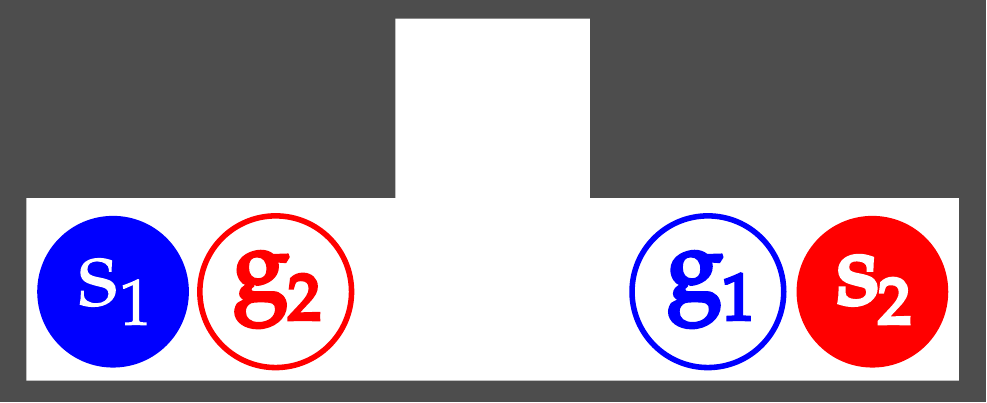}
\par\end{centering}

\caption{\label{fig:Instance_where_PP_fails}The picture shows two robots desiring
to move from $s_{1}$ to $g_{1}$ ($s_{2}$ to $g_{2}$ resp.) in
a corridor that is only slightly wider than a body of a single robot.
The scenario assumes that both robots have identical maximum speeds.
Observe that irrespective of which robot starts planning first, its
trajectory will be in conflict with all satisfying trajectories of
the robot that plans second.}
\end{figure}

\par\end{center}

Let us now analyze when is prioritized planning bound to fail. The
algorithm fails to find a trajectory for robot $i$ if 1) no satisfying
path exists for robot $i$, i.e. the robot cannot reach its destination
even if there are no other robots in the workspace; 2) every satisfying
trajectory of robot $i$ is in conflict with some higher-priority
robot. There are two types of conflicts that can occur between a satisfying
trajectory $\pi$ of robot $i$ and a higher-priority robot:

\textbf{Type~A}: Occurs if trajectory $\pi$ is in conflict with
a higher-priority robot who has reached and is ``sitting'' at its
destination, i.e. it is blocked by a static higher-priority robot.
The following Figure shows a scenario where all satisfying trajectories
of a robot are in Type A conflict:

\begin{center}
\begin{figure}[H]
\begin{centering}
\includegraphics[scale=0.5]{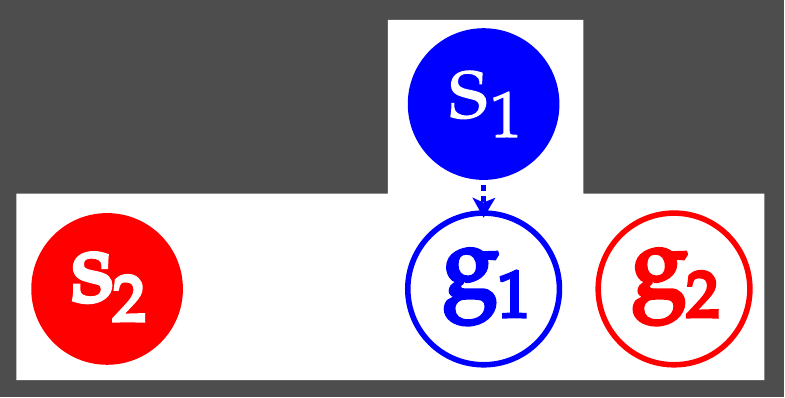}
\par\end{centering}

\caption{Robot 1 travels from $s_{1}$ to $g_{1}$, robot 2 travels from $s_{2}$
to $g_{2}$. Both robots have identical maximum speed $v=1$. Robot
1 plans first and adopts a straight line trajectory from $s_{1}$
to $g_{1}$ at the maximum speed, because it ignores the task of robot
2. Consequently all satisfying trajectories of robot 2 will be in
Type A conflict with robot 1.}
\end{figure}

\par\end{center}

\textbf{Type~B}: Occurs if trajectory $\pi$ of robot $i$ is in
conflict with a higher-priority robot who is moving towards its destination,
i.e. it is ``run over'' by a moving higher-priority robot. The following
Figure shows a scenario where all satisfying trajectories of a robot
are in Type B conflict:

\begin{center}
\begin{figure}[H]
\begin{centering}
\includegraphics[scale=0.5]{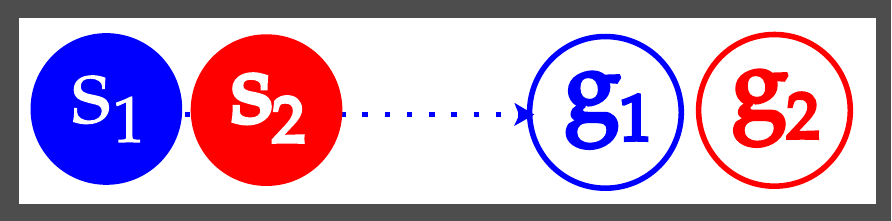}
\par\end{centering}

\caption{Robot 1 travels from $s_{1}$ to $g_{1}$, robot 2 travels from $s_{2}$
to $g_{2}$. Assume that robot 1 can travel \emph{twice as fast} as
robot 2. Then, robot 1 will plan first and adopts a straight line
trajectory from $s_{1}$ to $g_{1}$ at the maximum speed, because
it ignores the task of robot 2. Robot 2 plans second, but all satisfying
trajectories for robot 2 are in Type B conflict with robot 1. If no
satisfying trajectory exists or all satisfying trajectories are engaged
in a Type A or Type B conflict, then prioritized planing fails to
find a satisfying and consistent trajectory for robot $i$ and terminates
with failure. }
\end{figure}

\par\end{center}

A question that naturally arises is whether it would be possible
to restrict the class of solvable instances or to alter the prioritized
planning algorithm such that there will always be at least one trajectory
without neither Type A nor Type B conflict for each robot. 

One way to ensure that there will be a satisfying trajectory without
Type A conflict for every robot is to only consider instances, where
each robot has a path to its goal that avoids goal regions of all
higher-priority robots. When each robot follows such a path, then
they cannot be engaged in a Type A conflict, because a Type A conflict
can only occur at the goal region of one of the higher-priority robots.

Unfortunately, the existence of a trajectory without Type B conflict
is difficult to guarantee in classical prioritized planning, since
higher-priority robots completely ignore interactions with lower-priority
robots when planning their trajectories. To ensure that each robot
will have a satisfying trajectory without Type B conflict, all higher-priority
robots would have to plan their trajectories so that the lower-priority
robots are always left with some alternative trajectory that can be
used to avoid the potential conflicts of this type. 

One way to ensure that there will be a satisfying trajectory without
Type B conflict for every robot is to consider only instances where
each robot has a path to its goal that avoids start region of lower-priority
robots and enforce that the trajectory of each robot will avoid start
regions of all lower-priority robots. When this is ensured, then any
robot will always have a fall-back option to wait at its start position
(since no higher-priority robot can run over its start region) until
its desired path is clear of all higher-priority robots. Thus it can
always avoid Type B conflicts.

Moreover, if the robot continues by following a path that avoids goal
regions of higher-priority robots, then the resulting trajectory is
also guaranteed to avoid the Type A conflicts.
\begin{thm}
\label{thm:if-S-G-avoiding-paths-exist-then-seq-solution-exists}Let
us have a trajectory coordination problem with workspace $\mathcal{W}$
and tasks $\left\langle s_{1},g_{1}\right\rangle ,\ldots,\left\langle s_{n},g_{n}\right\rangle $
for robots $1,\ldots,n$. If for every robot $i$ there exist a $S^{>i}$-avoiding
and $G^{<i}$-avoiding satisfying path, then a sequential conflict-free
solution can be constructed.\end{thm}
\begin{IEEEproof}
We will construct the solution inductively as follows: 

\textbf{Induction assumption:} Trajectories of robots $1,\ldots,i-1$
are satisfying and $S^{>i-1}$-avoiding.

\textbf{Base step (robot $1$):} Robot $1$ is the highest-priority
robot. There are no higher-priority robots that the robot 1 needs
to avoid. From our assumption there exists a path $p$ that is satisfying
for robot $1$ and $S^{>1}$-avoiding. A satisfying and $S^{>1}$-avoiding
trajectory for robot 1 can be simply constructed by following the
path $p$ at an arbitrary positive speed. Such a trajectory can always
be constructed, therefore the algorithm will not report failure when
planning for robot $1$.

\textbf{Induction step (robot $i$): }From our assumption there exists
a path $p$ that is satisfying for robot $i$, $S^{>i}$-avoiding
and $G^{<i}$-avoiding. Since all trajectories for robots $1,\ldots,i-1$
are satisfying (i.e. eventually reach the goal and stay there) then
there must exists a time point $\overline{t}$ after which all robots
$1,\ldots,i-1$ have reached and will stay at their goal. A satisfying
and $S^{>i}$-avoiding trajectory for robot $i$ that is conflict-free
with all robots $1,\ldots i-1$ can be constructed as follows:
\begin{itemize}
\item In interval $[0,\overline{t}]$ stay at $s_{i}$. The trajectory cannot
be in conflict with higher-priority robots during this interval, because
all trajectories of robots $1,\ldots,i-1$ are $S^{>i-1}$ and thus
also $S^{i}$-avoiding. 
\item In interval $[\overline{t},\infty]$ follow path $p$ until the goal
position $g_{i}$ is reached. The path $p$ avoids regions $G^{<i}$
and thus the trajectory cannot be in collision with any of the higher-priority
robots $1,\ldots,i-1$ because they are at their goal positions during
this time interval, which the path $p$ avoids.
\end{itemize}
Such a trajectory can always be constructed. 

The trajectories of robots $1,\ldots,i-1$ are satisfying and $S^{>i-1}$-avoiding,
which implies that they are also $S^{>i}$-avoiding. The newly computed
trajectory for robot $i$ is satisfying and $S^{>i}$-avoiding. By
taking the union of the old set of trajectories and the new trajectory
we have a set of trajectories for robots $1,\ldots,i$ that are satisfying
and $S^{>i}$-avoiding.
\end{IEEEproof}
As we can see from the constructive proof of Theorem~\ref{thm:if-S-G-avoiding-paths-exist-then-seq-solution-exists},
the instances that admit $S^{>i}$-avoiding and $G^{<i}$-avoiding
paths for each robot can be solved by a simple sequential algorithm
that navigates all robots one-after-another along their shortest $S^{>i}$-avoiding
and $G^{<i}$-avoiding paths. Albeit simple, such an approach never
lets two robots to move concurrently and thus it typically generates
solutions of poor quality. The solution quality can be improved if
we adopt the prioritized planning approach and find for each robot
a best $S^{>i}$-avoiding trajectory that avoids conflicts with higher-priority
robots.

\subsection*{Revised Prioritized Planning}

We propose a Revised version of Prioritized Planning (RPP) that uses
the insights from the preceding discussion and plans the trajectory
of each robot so that both a) start position of all lower-priority
robots are avoided and b) conflicts with higher-priority robots are
avoided. The pseudocode of RPP is listed in Algorithm~\ref{alg:RPP}.

\SetKwFunction{rpp}{RPP}
\begin{algorithm}
\alg{\rpp}{

	$\dobst\leftarrow\emptyset$\;

	\For{$i\gets1\ldots n$}{

		$S\leftarrow$$\underset{j>i}{\bigcup}S^{j}$

		$\pi_{i}\gets$\besttraj{$\mathcal{W}\setminus S,\dobst$}\;\label{alg:rpp-best-traj}

		\If{$\pi_{i}=\emptyset$}{

		 report failure and terminate\;

		}

		$\dobst\leftarrow\dobst\cup R_{i}^{\Delta}(\pi_{i})$\;

		}

	}

\caption{\label{alg:RPP}Revised Prioritized Planning}
\end{algorithm}

\subsubsection*{Properties}

The RPP algorithm inherits the termination and soundness properties
from the PP algorithm. The algorithm terminates successfully in $n$
iterations if a trajectory for each robot has been found. The algorithm
terminates with failure at iteration $i<n$ if there is a robot $i$
for whom a satisfying trajectory in $\mathcal{W}\setminus S$ has
not been found. 

In general, it is not guaranteed that a trajectory that avoids both
start positions of lower-priority robots and regions occupied by higher-priority
robots will exist for each robot and thus the algorithm may fail to
provide a solution to a solvable problem instance. Consider e.g. the
example in Figure~\ref{fig:Instance_where_PP_fails} once again.
However, for the instances characterized by the following condition,
the solution is guaranteed exists and RPP will find it.
\begin{cor}
\label{cor:RPP-is-doesnt-fail-if-S-G-avoiding-paths-exist}If there
is a $S^{>i}$-avoiding, $G^{<i}$-avoiding satisfying path for every
robot $i$ and a complete algorithm is used for the single-robot trajectory
planning in \besttraj function, then RPP is guaranteed to terminate
with conflict-free solution.\end{cor}
\begin{IEEEproof}
Consider the inductive argument from the proof of Theorem~\ref{thm:if-S-G-avoiding-paths-exist-then-seq-solution-exists}.
The argument states that at every iteration, there exists a $S^{>i}$-avoiding
satisfying trajectory for robot $i$ that avoids all higher-priority
robots. Since the single-robot planning algorithm is assumed to be
complete, it cannot fail in finding such a trajectory.
\end{IEEEproof}

\subsection*{Valid Infrastructures}

Consider a situation when one designs a closed multi-robot systems
such as a warehouse with a large number of autonomous vehicles. Then,
we can exploit the Theorem~\ref{thm:if-S-G-avoiding-paths-exist-then-seq-solution-exists}
and Corollary~\ref{cor:RPP-is-doesnt-fail-if-S-G-avoiding-paths-exist}
to design the environment and allowed tasks of the robots in such
a way that $S^{>i}$-avoiding and $G^{<i}$-avoiding paths will always
exists and RPP will be consequently guaranteed to provide a conflict-free
solution to every trajectory coordination query. An important class
of environments that satisfy the condition of having a $S^{>i}$-avoiding
and $G^{<i}$-avoiding paths for every possible task of every robot
are \emph{valid infrastructures}.

Let $D(x,r)$ be a closed disk centered at $x$ with radius $r$ and
$\interior_{r}\, X$ be an $r$-interior of set $X$ defined as 
\[
\interior_{r}\, X:=\left\{ x:D(x,r)\subseteq X\right\} .
\]
Any path that lies entirely in $\interior_{r}\, X$ will have $r$-clearance
with respect to $X$, i.e. any point on the path will be at minimum
distance $r$ from the closest boundary of $X$. 

\begin{figure}
\begin{centering}
\subfloat[Valid infrastructure: The workspace $\mathcal{W}$ and endpoints $\{e_{1},e_{2},e_{3},e_{4}\}$
for robot having radius $r$ form a valid infrastructure. ]{\centering{}\includegraphics[width=0.4\columnwidth]{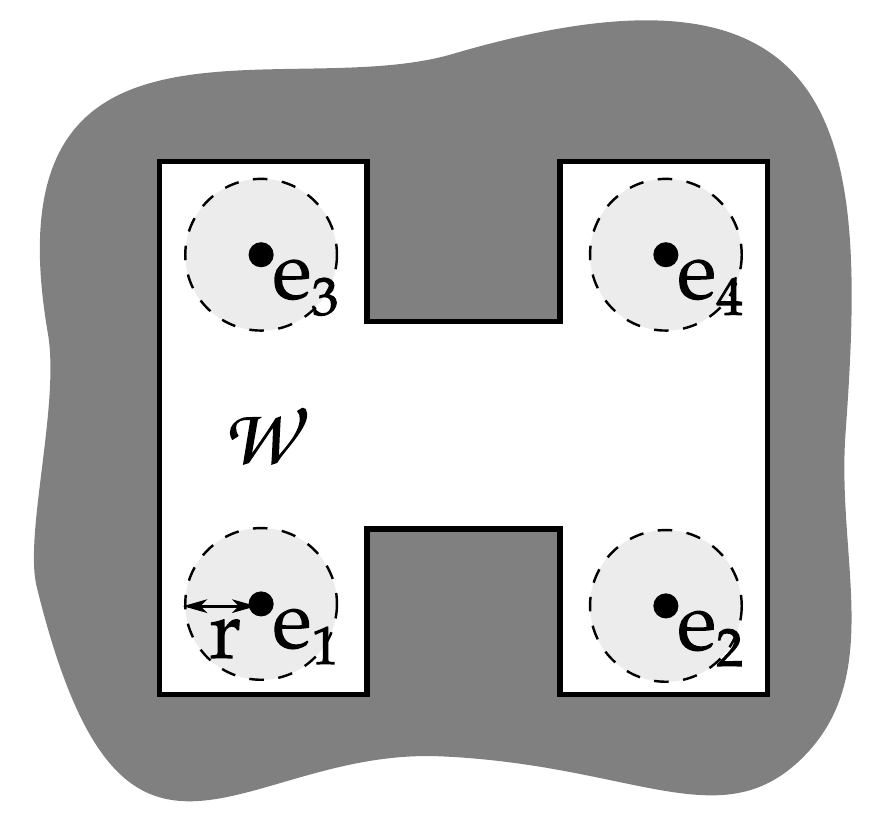}}~~~~\subfloat[Invalid infrastructure: The workspace $\mathcal{W}$ and endpoints
$\{e_{1},e_{2},e_{3}\}$ do not form a valid infrastructure because
there is no path from $e_{1}$ to $e_{2}$ with $2r$-clearance to
$e_{3}$ for a robot having radius $r$.]{\begin{centering}
\includegraphics[width=0.4\columnwidth]{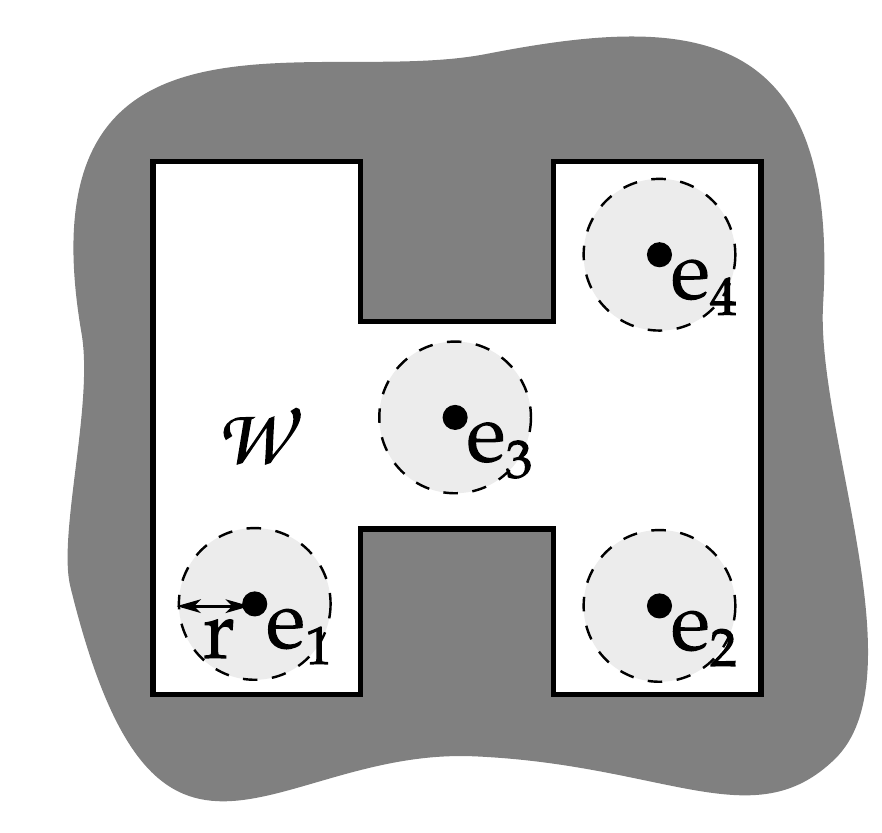}
\par\end{centering}

}
\par\end{centering}

\caption{\label{fig:infrastructure}Valid and invalid infrastructure }
\end{figure}

We say that a workspace $\mathcal{W}$ together with a set of endpoints
$E$ form a \emph{valid infrastructure} for circular robots with maximum
radius $r$ if any two endpoints can be connected by a path in workspace
$\interior_{r}\left(\mathcal{W}\setminus\underset{e\in E}{\cup}D(e,r)\right)$,
i.e. there must exists a path between any two endpoints with at least
$r$-clearance to the boundary of the workspace and at least $2r$-clearance
to any other endpoint. Figure~\ref{fig:infrastructure} illustrates
the concept of a valid infrastructure.

The notion of valid infrastructures follows the structure typically
witnessed in man-made environments that are intuitively designed to
allow efficient transit of multiple people or vehicles. In such environments,
the endpoint locations where people or vehicle stop for long time
are separated from the transit area that is reserved for travel between
these locations. 

In a road network, for example, the endpoints would be the parking
places and the system of roads is built in such a way that any two
parking places are reachable without crossing any other parking place.
Similar structure can be witnessed in offices or factories. The endpoints
would be all locations, where people may need to spend longer periods
of time, e.g. surroundings of the work desks or machines. As we know
from our every day experience, work desks and machines are typically
given enough free room around them so that a person working at a desk
or a machine does not obstruct people moving between other desks or
machines. We can see that real-world environments are indeed often
designed as valid infrastructures. 

Suppose that a particular workspace $\mathcal{W}$ and a set of endpoints
$E$ form a valid infrastructure for robots having radius $r$. An
instance of multi-robot trajectory coordination problem in such an
infrastructure would then involve a number of robots traveling from
one endpoint to another so that each endpoint is used by at most one
robot. From the valid infrastructure property, we know that for each
robot there is a path $p$ from its start endpoint to its goal endpoint
that avoids all other endpoints in the infrastructure with $2r$ clearance.
Since all other robots' start and goal positions lie at some endpoint,
the path $p$ is also avoiding start and goal position of every other
robot. In other words, for every robot there is a path that is $S^{-i}$-avoiding
and $G^{-i}$-avoiding, which implies that the path is also $S^{>i}$-avoiding
and $G^{<i}$-avoiding. Therefore, all trajectory coordination queries
between the endpoints of a valid infrastructure will be successfully
solved by RPP algorithm, given that a complete algorithm is used for
the single robot trajectory planning.

\subsection*{Checking Solvability}

Now consider an open multi-robot system in which new robot tasks can
appear any time and we cannot guarantee that the robots will move
only between pre-designed endpoints. Then we can use Theorem~\ref{thm:if-S-G-avoiding-paths-exist-then-seq-solution-exists}
to check whether adding a new robot task maintains sequential solvability
of the trajectory coordination problem between the existing robots
and the new robot. If it does not, then we can decide to reject or
delay the adding of new robot task. Corollary~\ref{cor:RPP-is-doesnt-fail-if-S-G-avoiding-paths-exist}
gives us sufficient condition that can be used to quickly determine
whether a particular problem instance is solvable by RPP without actually
running the algorithm. This is done by verifying that $S^{>i}$-avoiding
and $G^{<i}$-avoiding satisfying path exists for each robot. Deciding
whether such paths exist amounts to planning $n$ spatial paths amidst
static obstacles, which is in practice significantly faster than planning
$n$ spatio-temporal trajectories amidst dynamic obstacles which would
have to be done if RPP is executed.

\subsection*{Limitations}

We have shown that there is a class of instances that RPP completely
covers, but PP does not. However, outside this class we can find instances
that PP solves, but RPP does not. Consider the following scenario:

\begin{center}
\begin{figure}[H]
\begin{centering}
\includegraphics[scale=0.5]{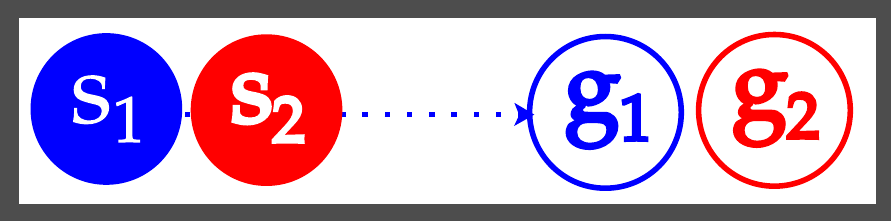}
\par\end{centering}

\caption{Robot 1 travels from $s_{1}$ to $g_{1}$, robot 2 travels from $s_{2}$
to $g_{2}$. Assume that both robots can travel at the same maximum
speed. Robot 1 searches for a trajectory that avoids start position
of robot $2$; such a trajectory does not exist and thus RPP finishes
with failure. Note that PP will successfully find a solution to this
instance: Robot 1 will plan a trajectory that follows straight line
at maximum speed. Robot 2 will search for a trajectory that avoids
the trajectory of robot 1 and finds that it suffices to travel at
maximum speed to its goal $g_{2}$.}
\end{figure}

\par\end{center}

None of the algorithm is therefore superior to the other in terms
of instance coverage. 

Further, since RPP avoid start regions preemptively, even when they
can be safely passed through, the solutions generated by RPP tend
to be be slightly longer than the ones generated by PP. This can be
demonstrated in the following scenario:

\begin{center}
\begin{figure}[H]
\begin{centering}
\bigskip{}
\includegraphics[scale=0.5]{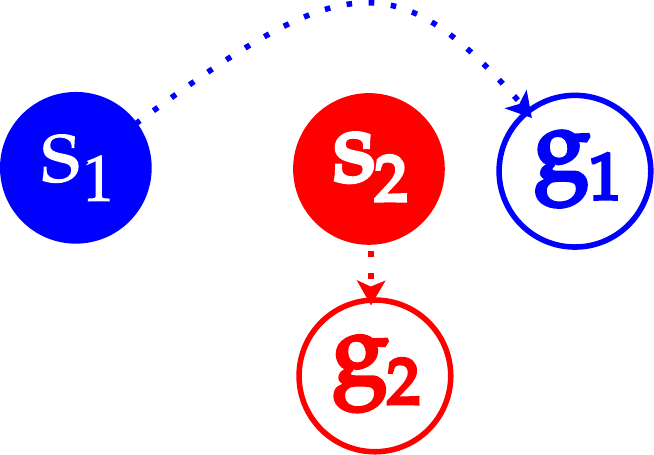}
\par\end{centering}

\caption{\label{fig:RPP-returns-worse-solution-than-PP}When RPP searches for
a trajectory for robot $1$ it has to avoid start position of robot
$2$, resulting in the curved trajectory as depicted in the picture.
On the other hand, PP would generate a shorter straight-line trajectory
connecting start and destination of robot $1$.}
\end{figure}

\par\end{center}

We can see that despite the theoretical guarantees of RPP, there exist
situations in which PP would be a more appropriate choice than RPP.

\section{Decentralized Algorithms}

Imagine a multi-robot system consisting of a large number of heterogeneous
autonomous robots. In such a scenario, a decentralized implementation
of (revised) prioritized planning may be more desirable than a centralized
one. In a decentralized implementation, each robot runs its own instance
of the algorithm and exchanges messages with the other robots according
to a prescribed communication protocol. If an inconsistency is detected
by a robot, then it recomputes the best trajectory for itself using
its own on-board computation resources. The process should eventually
converge to a state where all robots hold mutually conflict-free trajectories. 

An advantage of such an approach is that several robots often end
up computing their trajectories in parallel and thus a conflict-free
solution is usually computed faster. Another advantage for multi-robot
systems with heterogeneous robots is that the kinematic and other
potentially implicit constraints on the trajectory of a particular
robot stay local to that robot and do not need to be formalized nor
communicated, which simplifies the design of the communication protocol
and allows each robot to use a custom robot-specific planner for planning
its trajectory.

\subsection*{\label{sub:Decentralized,-but-Synchronized}Synchronized Decentralized
Implementation}

A decentralized implementation of classical prioritized planning scheme,
where robots concurrently proceed in synchronize rounds, has been
first presented by Velagapudi et al. in \cite{VelagapudiSS10}. We
will use their approach as a baseline decentralized implementation
of (revised) prioritized planning and denote the resulting algorithm
as \emph{synchronized decentralized implementation of (revised) prioritized
planning,} SD-(R)PP. 

The algorithm proceeds in synchronized rounds. During every round,
each robot ensures that its current trajectory is consistent with
the trajectories of higher-priority robots from the previous round.
If the current trajectory is consistent, then the robot keeps its
current trajectory and remains silent. Otherwise, it finds a new consistent
trajectory for itself and broadcasts the trajectory to all other robots.
When a robot finishes its computation in the current round, then it
waits for all other robots to finish the round and all robots simultaneously
proceed to the next round. The algorithm successfully finishes if
none of the robots changes its current trajectory during a single
round. The SD-PP algorithm finishes with failure if there is a robot
that fails to find a trajectory that avoids the higher-priority robots
following their respective trajectories. The SD-RPP algorithm, on
the other hand, finishes with failure if there is a robot that fails
to find a satisfying trajectory that avoids the start positions of
lower-priority robots. The pseudocode of SD-(R)PP is listed in Algorithm~\ref{alg:SDPP}.

\SetKw{var}{var}
\SetKw{broadcast}{broadcast}
\SetKw{Globals}{Global variables:}
\SetKwFunction{replan}{Replan}
\SetKwFunction{checkconsistency}{Check-consistency}
\SetKwFunction{assertconsistency}{Find-consistent}
\SetKwFunction{besttraji}{Best-traj$_i$}
\SetKwFunction{processmessages}{Process-messages}{}
\SetKwProg{handlemessage}{Handle-message}{}{} 
\SetKwFunction{sdrpp}{SD-(R)PP} 

\noindent 
\begin{algorithm}
\alg{\sdrpp}{

	$\pi_{i}\leftarrow\emptyset$\;

	$\ts{}_{i}\leftarrow\emptyset$\;

	$S\leftarrow\begin{cases}
\emptyset & \text{for SD-PP}\\
\underset{j>i}{\bigcup}S^{j} & \text{for SD-RPP}
\end{cases}$~\;

	\Repeat{ not global termination detected }{

		$\pi^{*}\leftarrow$ \assertconsistency{$\pi_{i},\mathcal{\mathcal{W}}\setminus S,\dobst(\ts_{i})$}\;

		\uIf{$\pi^{*}=\emptyset$}{

			 report failure and terminate\;

		}

		\ElseIf{$\pi^{*}\neq\pi_{i}$}{

			$\pi_{i}\leftarrow\pi^{*}$\;

			\broadcast$\mathrm{INFORM}(i,R_{i}^{\Delta}(\pi^{*}))$\;\label{alg:sdpp-broadcast}

		}

		 \textbf{wait} for INFORM messages from all other robots, wait
for all other robots to finish processing INFORM messages \;

	}

}

\handlemessage{$\mathrm{INFORM}(j,\Delta_{j})$}{\label{alg:sdpp-handlemessage-start}

	\If{$j<i$}{

	$\ts_{i}\leftarrow\left(\ts_{i}\setminus\left\{ (j,\Delta'_{j})\,:\,(j,\Delta'_{j})\in\ts_{i}\right\} \right)\cup\left\{ (j,\Delta_{j})\right\} $\;

	}

}

\function{\assertconsistency{$\pi,\mathcal{\mathcal{W}},\Delta$}}{

	\eIf{$\pi=\emptyset\,\vee\,\neg\mathrm{consistent}_{i}(\pi,\Delta)$}{\label{alg:sdpp-assert-consistency-condition}

		 $\pi^{*}\gets$\besttraji{$\mathcal{\mathcal{W}},\Delta$}\;\label{alg:sdpp-assert-consistency-best-traj}

		\Return$\pi^{*}$\; \label{alg:sdpp-assert-consistency-return-new-traj}

	} { 

		\Return $\pi$\;\label{alg:sdpp-assert-consistency-return-prev-traj}

	}

}

\caption{\label{alg:SDPP}Synchronized Decentralized Implementation of (Revised)
Prioritized Planning. Pseudocode for robot~$i$}

\end{algorithm}

In SD-(R)PP, each robot $i$ maintains a database of space-time regions
occupied by higher-priority robots. We call such a database a \trajectorystore{}
and model it as a set of pairs $\ts_{i}=\left\{ (j,\Delta_{j})\right\} $,
where $\Delta_{j}$ is the space-time region occupied by robot $j$.

Function $\dobst(\ts)$ represents the region of the space-time occupied
by all robots stored in a \trajectorystore{} $\ts$:

\[
\dobst(\ts):=\underset{(j,\Delta_{j})\in\ts}{\cup}\Delta_{j}.
\]

Further, we use a predicate $\mathrm{consistent_{i}}(\pi,\dobst)$
to express that robot $i$ following the trajectory $\pi$ is collision-free
against dynamic obstacles $\dobst$, defined as:
\begin{multline*}
\mathrm{consistent}_{i}(\pi,\dobst):=R_{i}^{\Delta}(\pi)\cap\dobst=\emptyset.
\end{multline*}

\subsubsection*{Properties}

In order to facilitate and simplify exposition of the later introduced
asynchronous algorithm, we developed an alternative proof of termination
of the SD-(R)PP algorithm, which deviates from the original one devised
by the authors of SD-PP in~\cite{VelagapudiSS10}. Further, in this
section we show that SD-(R)PP inherits the soundness and completeness
properties from its respective centralized counterpart.

The SD-(R)PP algorithm is guaranteed to terminate. First we need to
define what does termination mean for a decentralized algorithm. A
decentralized algorithm
\begin{itemize}
\item \textbf{terminates} when all robots stop computing,
\item \textbf{terminates with failure} if it terminates and there is at
least one robot that reported failure during the computation,
\item \textbf{terminates successfully} if it terminates and does not terminate
with failure.
\end{itemize}
To show that SD-(R)PP terminates, we first show that robots running
SD-(R)PP cannot exchange messages forever. 
\begin{prop}
\label{prop:All-robots-running-SDRPP-stop-sending-messages}All robots
running SD-(R)PP algorithm eventually stop sending INFORM messages.\end{prop}
\begin{IEEEproof}
We will proceed by induction on the robot priority~$i$.\\
\textbf{Inductive hypothesis:} Robots $1,\ldots,i-1$ eventually stop
sending messages.\textbf{}\\
\textbf{Base step} (robot 1):\\
Robot 1 is the highest-priority robot and as such it does not receive
any message from a higher-priority robot. Therefore, its \trajectorystore{}
will stay empty. During the initialization, the robot $1$ either
successfully finds its initial trajectory and broadcasts a single
message or reports a failure and terminates.\textbf{ }Since its \trajectorystore{}
is empty, its initial trajectory will never become inconsistent, the
robot will therefore never replan and send any further INFORM message.
\textbf{}\\
\textbf{Induction step} (robot $i$): \\
From the inductive hypothesis we know that each of the robots $1,\ldots,i-1$
eventually stops broadcasting messages. After the last message from
the robots $1,\ldots,i-1$ has been received by robot $i$, its \trajectorystore{}
gets updated for the last time, since from our assumption there are
no more messages from higher-priority robots. After \trajectorystore{}
changes for the last time, the robot either a) keeps its current trajectory
if it is consistent with the last \trajectorystore{}, b) finds a
new consistent trajectory or c) terminates with failure. In cases
a) and b), the current trajectory will never become inconsistent again
because \trajectorystore{} does not change anymore and thus robot
will never have to replan and communicate the new trajectory. In case
c), the robot has terminated and thus it won't send any further messages.
We see that after robots $1,\ldots,i-1$ stopped sending messages,
also robot $i$ eventually stops sending messages.\end{IEEEproof}
\begin{cor}
SD-(R)PP terminates.\end{cor}
\begin{IEEEproof}
We know that all robots in the system will eventually stop sending
messages. Assume that the last message is broadcast during round $k$.
In the round $k+1$, no robot changes its trajectory since otherwise
a message would have to be broadcast which is a contradiction. If
no robot changes its trajectory during the round, the global termination
condition is satisfied and the system terminates.
\end{IEEEproof}
Unless a failure is reported by one of the robots, the solution computed
when SD-(R)PP terminates is sound:
\begin{prop}
\label{thm:SD-RPP-is-sound}When SD-(R)PP successfully terminates,
then all robots hold trajectories that are mutually conflict-free.\end{prop}
\begin{IEEEproof}
Let $\pi_{i}$ be the trajectory of robot $i$ after the algorithm
terminated. We need to show that 
\[
\forall i,j:\; i\neq j\Rightarrow\pi_{i}\text{ and }\pi_{j}\text{ are conflict-free}.
\]
Take two arbitrary, but different robots $i$,$j$. Since the conflict-free
relation is symmetrical, we can assume $j<i$ w.l.o.g.  If robot $i$
stopped computing without failure, then it must have received the
INFORM message from higher-priority robot $j$ carrying its last trajectory
$\pi_{j}$ at some point before its termination. Since there are no
further INFORM messages broadcast by robot $j$, the trajectory store
of robot $i$ will contain $\pi_{j}$ from that point on. Every trajectory
returned by \assertconsistency function for robot $i$ from that
point on will be conflict-free with $\pi_{j}$ and thus also its last
trajectory $\pi_{i}$ will be conflict-free with $\pi_{j}$.
\end{IEEEproof}
The synchronized decentralized implementation of PP and RPP inherit
completeness properties from the respective centralized implementations.
In general, both SD-PP and SD-RPP are incomplete. However if $S^{>i}$-avoiding
and $G^{<i}$-avoiding satisfying path exists for each robot, then
the SD-RPP is guaranteed to terminate successfully. 
\begin{prop}
\label{thm:SD-RPP-solves-instances-with-S-G-avoiding-paths}If there
is a $S^{>i}$-avoiding, $G^{<i}$-avoiding satisfying path for every
robot $i$ and a complete algorithm is used for the single-robot trajectory
planning in \besttraj function, then SD-RPP is guaranteed to terminate
with a conflict-free solution.\end{prop}
\begin{IEEEproof}
The argument used in the proof of Theorem~\ref{cor:RPP-is-doesnt-fail-if-S-G-avoiding-paths-exist},
which shows that RPP will never fail during planning, can be extended
to decentralized implementations of RPP: Take arbitrary replanning
request for robot $i$. All trajectories of each higher-priority robot
$j<i$ have been generated to be $S^{>j}$-avoiding and thus such
trajectory will be also $S^{i}$-avoiding. All trajectories in the
trajectory store of robot $i$ are therefore $S^{i}$-avoiding. An
$S^{>i}$-avoiding satisfying trajectory consistent with trajectories
of higher-priority robots can be constructed as follows. Wait at start
position $s_{i}$ until all higher-priority robots reach their goal
position and then follow the $S^{>i}$-avoiding $G^{<i}$-avoiding
satisfying path from the assumption. Since such a trajectory is guaranteed
to exist for robot $i$ and a complete replanning algorithm is used,
the replanning cannot report failure. The algorithm must terminate
with success.
\end{IEEEproof}

\subsection*{\label{sec:Asynchronous-Prioritized-Planning}Asynchronous Decentralized
Implementation}

\begin{figure*}
\begin{centering}
\includegraphics[scale=0.5]{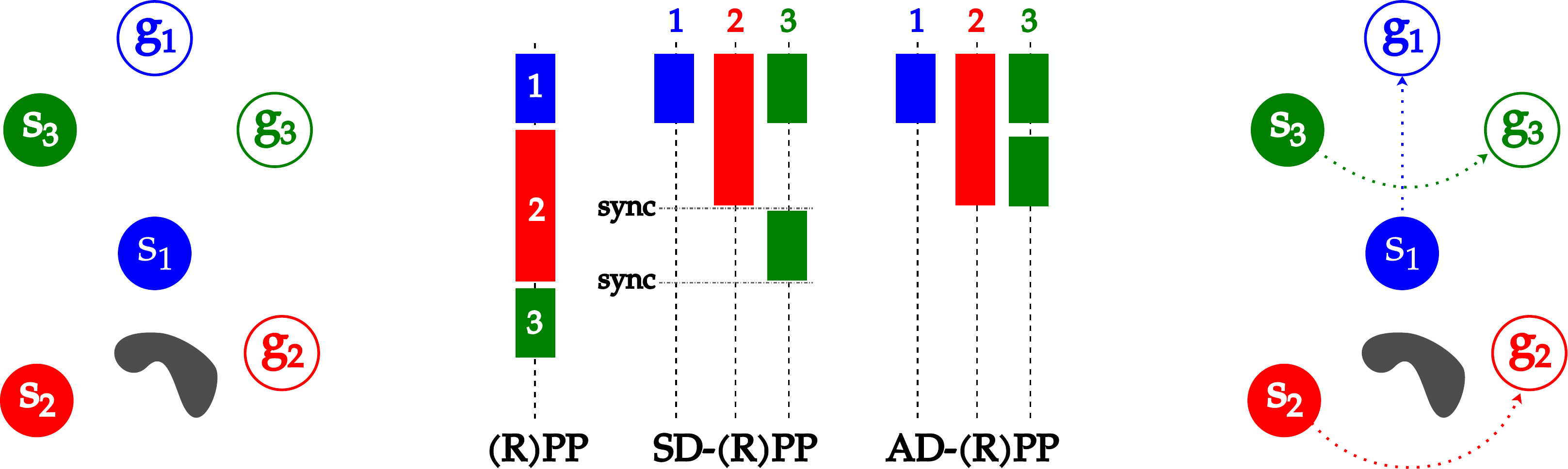}\\
\bigskip{}

\par\end{centering}

\caption{\label{fig:Example}Example problem in which AD-(R)PP converges faster
than SD-(R)PP. \textbf{Left:} The task of robots 1, 2, and 3. \textbf{Middle:}
Sequence diagrams showing the planning process in (R)PP, SD-(R)PP
and AD-(R)PP. Suppose that robot 2 needs to plan longer, because it
has to plan around the gray obstacle. In SD-(R)PP, the robot 3 starts
resolving the conflict with robot 1 only when robot 2 has finished
computing its trajectory in the first round. In AD-(R)PP, robot 3
stars resolving the conflict with robot 1 immediately after it becomes
aware of it, therefore it finds the solution faster. \textbf{Right:}
The final solution.}
\end{figure*}

Due to its synchronous nature, the SD-(R)PP algorithm  does not fully
exploit the computational resources distributed among individual robots.
In every iteration, the robots that finished their trajectory planning
routine sooner, or did not have to re-plan at all, idle while waiting
for the slower computing robots in that round, even though they could
use the time to resolve some of the conflicts they have among themselves
and speed up the overall process. An example of a situation, where
the asynchronous algorithm would be beneficial is illustrated in Figure~\ref{fig:Example}.

To deal with such an inefficiency, we propose an asynchronous decentralized
implementation of the (revised) prioritized planning scheme, abbreviated
as AD-(R)PP. The pseudocode code of AD-(R)PP is exposed in Algorithm~\ref{alg:ADPP}.
The asynchronous algorithm replaces the concept of globally synchronized
rounds (while loop in Algorithm~\ref{alg:SDPP}) by a reactive approach
in which every robot reacts merely to incoming \noun{INFORM} messages.
Upon receiving an \noun{INFORM} message \noun{(}\textbf{Handle-message}
$\mathrm{INFORM}(j,\Delta_{j})$ routine in Algorithm~\ref{alg:ADPP}),
the robot simply replaces the information about the trajectory of
the sender robot in its \trajectorystore{} and checks whether its
current trajectory is still consistent with the new contents of its
\trajectorystore{}. If the current trajectory is inconsistent, the
robot triggers replanning and inform other robots about its new trajectory,
otherwise the robot keeps its current trajectory and remains silent.

\SetKwFunction{adrpp}{AD-(R)PP}
\begin{algorithm}
\alg{\adrpp}{

	$\pi_{i}\leftarrow\emptyset$\;

	$\ts_{i}\leftarrow\emptyset$\;

	$S\leftarrow\begin{cases}
\emptyset & \text{for AD-PP}\\
\underset{j>i}{\bigcup}S^{j} & \text{for AD-RPP}
\end{cases}$~\;

	$\pi_{i}\leftarrow$\assertconsistency{$\pi_{i},\mathcal{W\setminus}S,\dobst(\ts_{i})$}\;\label{alg:adpp-initial-assert-consistency}

	\eIf{$\pi_{i}=\emptyset$}{

		 report failure and terminate\;

	}{

		\broadcast$\mathrm{INFORM}(i,R_{i}^{\Delta}(\pi_{i}))$\;

		 \textbf{wait} for global termination\;

	}

}

\handlemessage{$\mathrm{INFORM}(j,\Delta_{j})$}{\label{alg:adpp-handlemessage-start}

	\If{$j<i$}{

	$\ts_{i}\leftarrow\left(\ts_{i}\setminus\left\{ (j,\Delta'_{j})\,:\,(j,\Delta'_{j})\in\ts_{i}\right\} \right)\cup\left\{ (j,\Delta_{j})\right\} $\;\label{alg:adpp-update-Pi-by-msg}

	$\pi^{*}\leftarrow$ \assertconsistency{$\pi_{i},\mathcal{\mathcal{W}}\setminus S,\dobst(\ts_{i})$}\;\label{alg:adpp-assert-consistency}

	\uIf{$\pi^{*}=\emptyset$}{

		 report failure and terminate\;

		}

	\ElseIf{$\pi^{*}\neq\pi_{i}$}{

			$\pi_{i}\leftarrow\pi^{*}$\;

		\broadcast$\mathrm{INFORM}(i,R_{i}^{\Delta}(\pi^{*}))$\;

	}

	}

}\label{alg:adpp-handle-message-end}

\caption{\label{alg:ADPP}Asynchronous Decentralized Implementation of (Revised)
Prioritized Planning}

\end{algorithm}

\subsubsection*{Properties}

AD-(R)PP inherits all the desirable properties from its synchronized
and centralized coutnerparts, i.e. it terminates and if it terminates
with success than all the robots will hold conflict-free trajectories.
Further, AD-RPP is guaranteed to solve instances that admit $S^{>i}$-avoiding
and $G^{<i}$-avoiding path for each robot.
\begin{prop}
AD-(R)PP terminates.\end{prop}
\begin{IEEEproof}
Recall that the inductive argument demonstrating that robots running
SD-(R)PP will eventually stop sending messages (Lemma~\ref{prop:All-robots-running-SDRPP-stop-sending-messages})
does not make use of the synchronization points in SD-(R)PP and thus
it is also valid for AD-(R)PP. By this argument, we know that there
is a finite number of messages being sent. We can observe that a robot
running AD-(R)PP performs computation only during initialization or
when it processes an incoming message. When all robots process their
last incoming messages, the system terminates.\end{IEEEproof}
\begin{prop}
When AD-(R)PP successfully terminates, then all robots hold trajectories
that are mutually conflict-free.\end{prop}
\begin{IEEEproof}
The proof of soundness of SD-(R)PP (Proposition~\ref{thm:SD-RPP-is-sound})
is directly applicable also to AD-(R)PP.\end{IEEEproof}
\begin{prop}
If $S^{>i}$-avoiding, $G^{<i}$-avoiding satisfying path exists for
every robot $i$ and a complete algorithm is used for the single-robot
trajectory planning in \besttraj function, then AD-RPP terminates. \end{prop}
\begin{IEEEproof}
The proof of Proposition~\ref{thm:SD-RPP-solves-instances-with-S-G-avoiding-paths},
where this property is demonstrated to hold for SD-RPP, is directly
applicable also for AD-RPP. 
\end{IEEEproof}

\section{Experimental Analysis}

We compare the performance of PP, RPP, SD-PP, SD-RPP, AD-PP, AD-RPP
in terms of coverage, runtime, communication complexity and solution
quality. The comparison was performed in three real-world environments
(see Figure~\ref{fig:Empty-hall-roadmap}, \ref{fig:Corridor} and
\ref{fig:Warehouse-docks}). For each of the environments we generated
two sets of problem instances: 1) In \emph{free-formed tasks} instance
set, each robot is assigned a task to move from a randomly selected
start position to a randomly selected goal position. 2) In \emph{infrastructure
tasks }instance set, we generated a set of endpoints that together
with a particular roadmap discretization of the environment form a
valid infrastructure; each robot is then assigned a random endpoint
as a start position and a randomly chosen endpoint as a destination
position.

For the decentralized algorithms (SD-PP, SD-RPP, AD-PP and AD-RPP),
we assume that each robot uses its own on-board CPU to compute its
trajectory. To measure the runtime characteristics of the execution
of decentralized algorithms, we emulate the concurrent execution of
the algorithms using a discrete-event simulation. The simulation measures
the execution time of each message handling and uses the information
to simulate the concurrent execution of the decentralized algorithm
as if it is executed on $n$ independent CPUs, where $n$ is the number
of robots. The concurrent process execution simulator was implemented
using Alite multi-agent simulation toolkit. The source code of the
entire experimental setup can be downloaded at http://agents.cz/\textasciitilde{}cap/adpp/.

All compared algorithms use identical best trajectory planner. The
best trajectory for each robots is obtained by searching a roadmap
extended with a discretized time-dimension using A{*} algorithm, where
the heuristic is the shortest path on the graph from the given node
to the goal node when the dynamic obstacles are ignored. 

The experiments have been performed on AMD Opteron 8356 2.3GHz, 8
GB RAM. For each algorithm we measure the following characteristics:

\paragraph*{Coverage}

We waited before each algorithm returns either success or failure
and counted the number of instances each of the algorithm successfully
solved. 

The following characteristics were measured only on instances that
were solved by all compared algorithms:

\paragraph*{Time to solution}

We measured the wall-clock runtime needed to compute a solution. For
the centralized planner we recorded the time of termination of the
centralized planner. For the decentralized algorithms we recorded
the time when the last robot detected global termination of the computation.

\paragraph*{Speed-up}

In order to be able to easier judge the effect of asynchronous execution
in AD-(R)PP algorithm we also compute the speed-up ratio for both
decentralized algorithms over their centralized counterparts. The
speed-up for algorithm $A$ on instance $i$ is computed as 
\begin{multline*}
\frac{\text{runtime of centralized variant of alg. \ensuremath{A}on instance }i}{\text{runtime of alg. }A\text{ on instance }i}\,,
\end{multline*}

where the centralized variant of AD-PP and SD-PP is PP, and the centralized
variant of AD-RPP and SD-RPP is RPP.

\paragraph*{Replannings/Communication}

Every time a robot running a decentralized algorithm adopts a new
trajectory (replans), the trajectory is broadcast to all other robots.
Therefore the number of replannings directly corresponds to the number
of INFORM message broadcast in the system.

\paragraph*{Prolongation}

The objective criterion we minimize is the sum of goal arrival times
for each robot. We measure the duration of the trajectory for each
robot and compute the prolongation coefficient for each instance as
\begin{equation}
\text{prolongation of alg. }A\text{ on instance }i=\frac{\sum_{i=1}^{n}t_{i}^{A}-t'_{i}}{\sum_{i=1}^{n}t'_{i}},\label{eq:cost}
\end{equation}
where $t_{i}^{A}$ is the time robot $i$ needs to reach its goal
position when it follows the trajectory computed by the algorithm
$A$ and $t'_{i}$ is the time the robot $i$ would have needed to
reach its goal following the shortest path on the roadmap if the collisions
with other robots were ignored.

\subsubsection*{Comparison with reactive planning}

In order to evaluate the practical advantages of the proposed planning
approaches, we also compare their coverage with the popular reactive
technique ORCA in our environments. ORCA is a control approach that
continuously observes positions and velocities of other robots in
a defined neighborhood. Should any potential collision be detected,
a linear program is used to compute a new collision averting velocity
that the robot should follow. If there are no eminent collisions,
the robot follows its preferred preferred velocity. In our implementation
the preferred velocity points at the globally shortest path from the
robots current position to the goal.

\subsection*{Environments}

\subsubsection*{Empty-hall environment}

\begin{figure}
\begin{centering}
\subfloat[\emph{\label{fig:Empty-hall-roadmap}Empty-hall }Environment. The
roadmap used for planning is depicted in gray. Infrastructure endpoints
are shown in red.]{\centering{}\includegraphics[width=0.9\columnwidth]{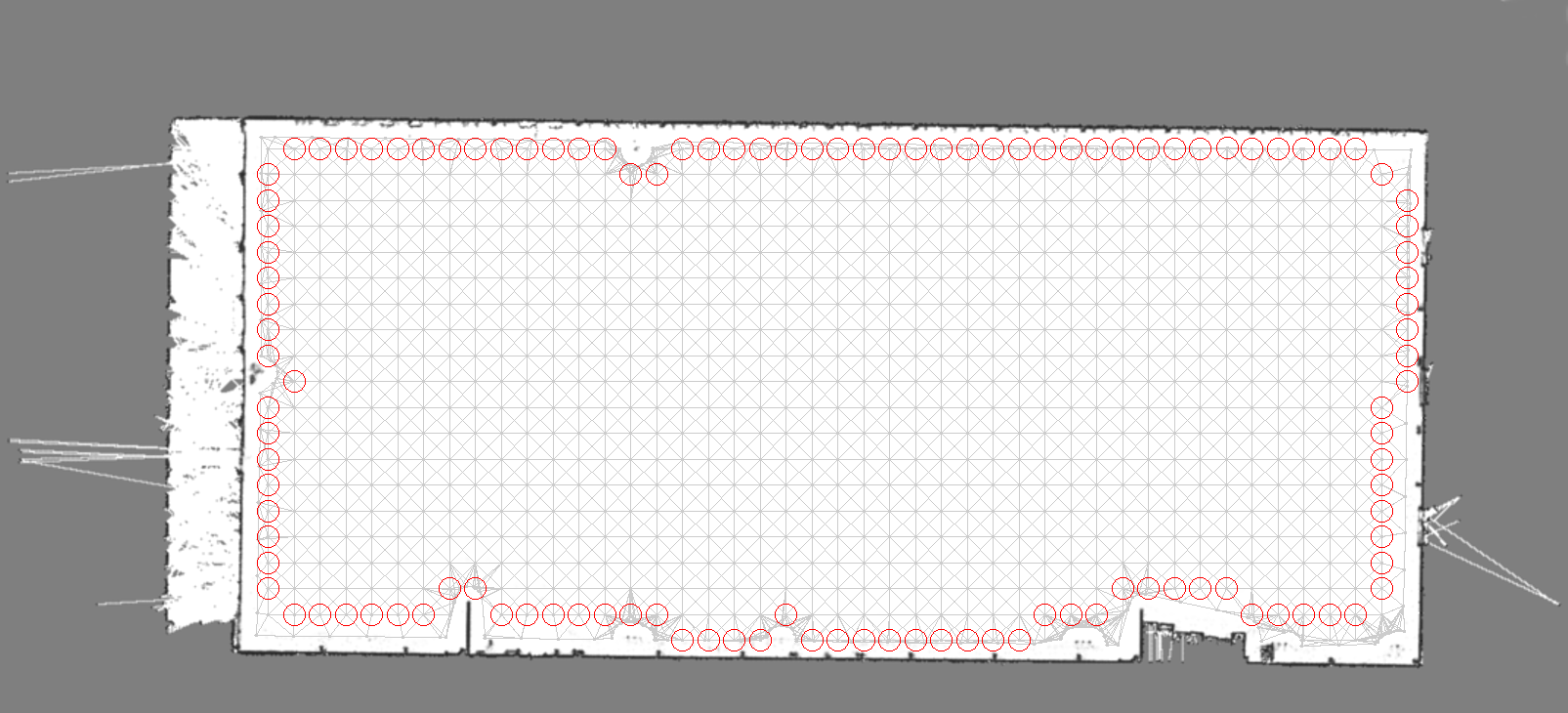}}
\par\end{centering}

\begin{centering}
\subfloat[Example free-formed tasks for 25 robots. Task of each robot shown
in blue.]{\begin{centering}
\includegraphics[width=0.9\columnwidth]{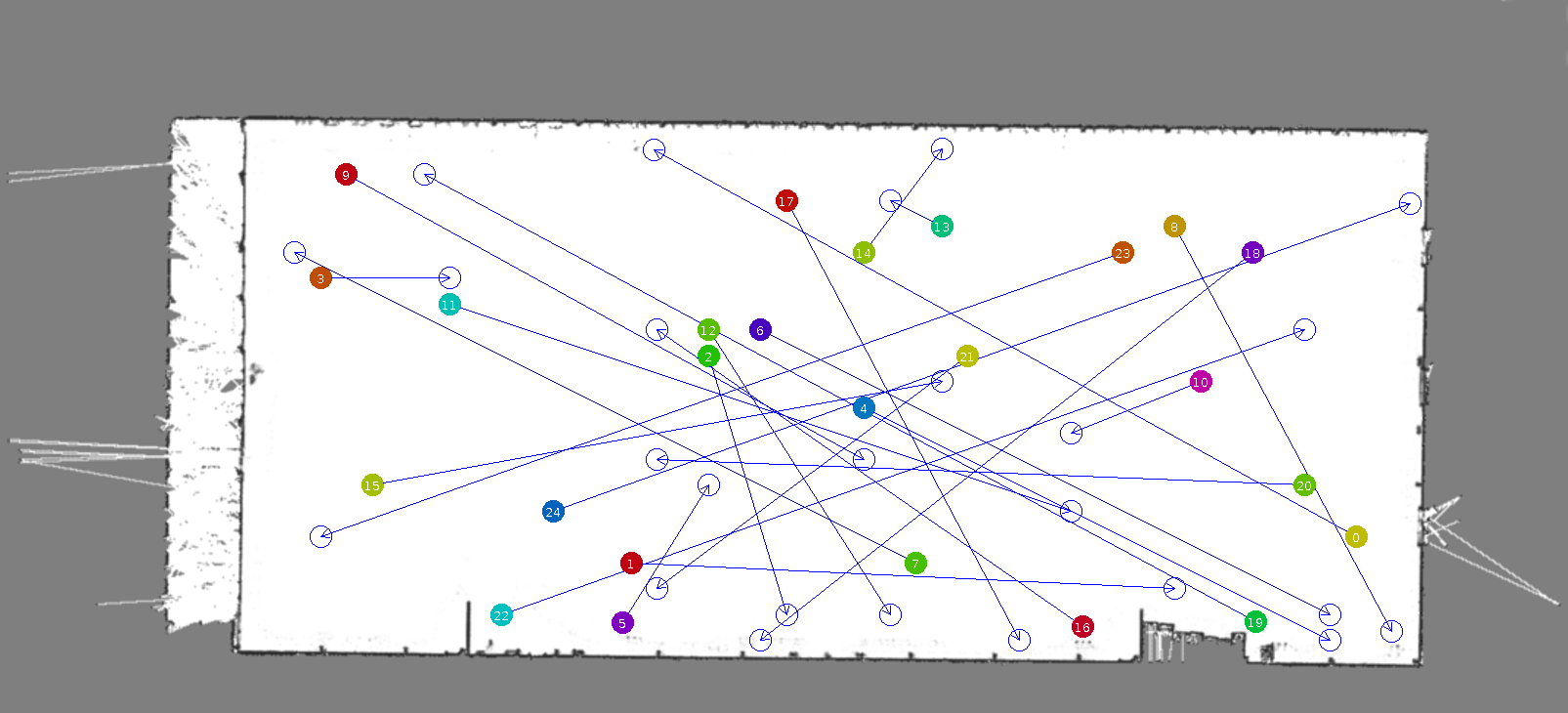}
\par\end{centering}

}
\par\end{centering}

\begin{centering}
\subfloat[Example infrastructure tasks for 25 robots. Task of each robot shown
in blue.]{\begin{centering}
\includegraphics[width=0.9\columnwidth]{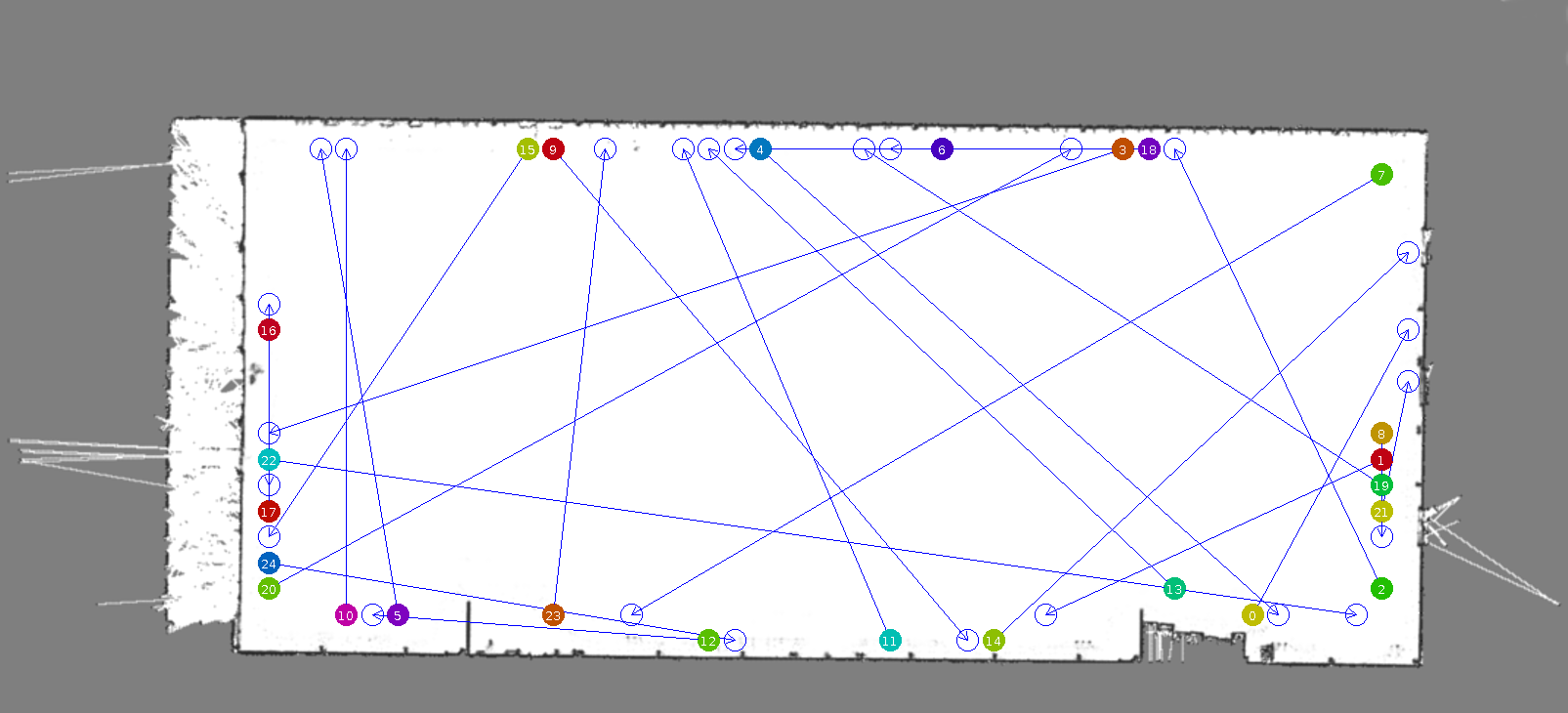}
\par\end{centering}

}
\par\end{centering}

\caption{\label{fig:Empty-hall-free}Empty hall environment}
\end{figure}

The \emph{empty-hall} environment and the roadmap used for trajectory
planning in the environment are depicted in Figure~\ref{fig:Empty-hall-roadmap}.
For both instance sets and each number of robots ranging from $n=1$
to $n=50$, we generated 25 random instances of the problem containing
the given number of robots.

\subsubsection*{Office Corridor environment}

\begin{figure}
\begin{centering}
\subfloat[\label{fig:Corridor} \emph{Office Corridor} Environment. The roadmap
used for planning is depicted in gray. Infrastructure endpoints are
shown in red.]{\centering{}\includegraphics[width=0.9\columnwidth]{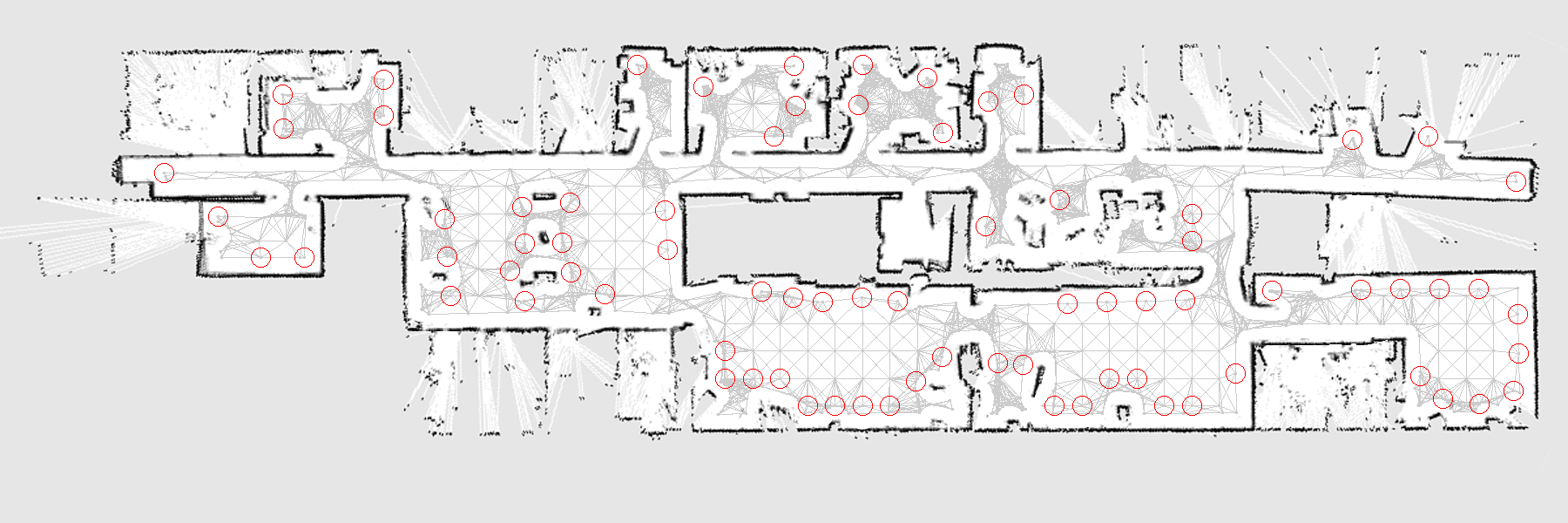}}
\par\end{centering}

\begin{centering}
\subfloat[\label{fig:Box-roadmap-2-1}Example free-formed tasks for 25 robots.
Task of each robot shown in blue.]{\begin{centering}
\includegraphics[width=0.9\columnwidth]{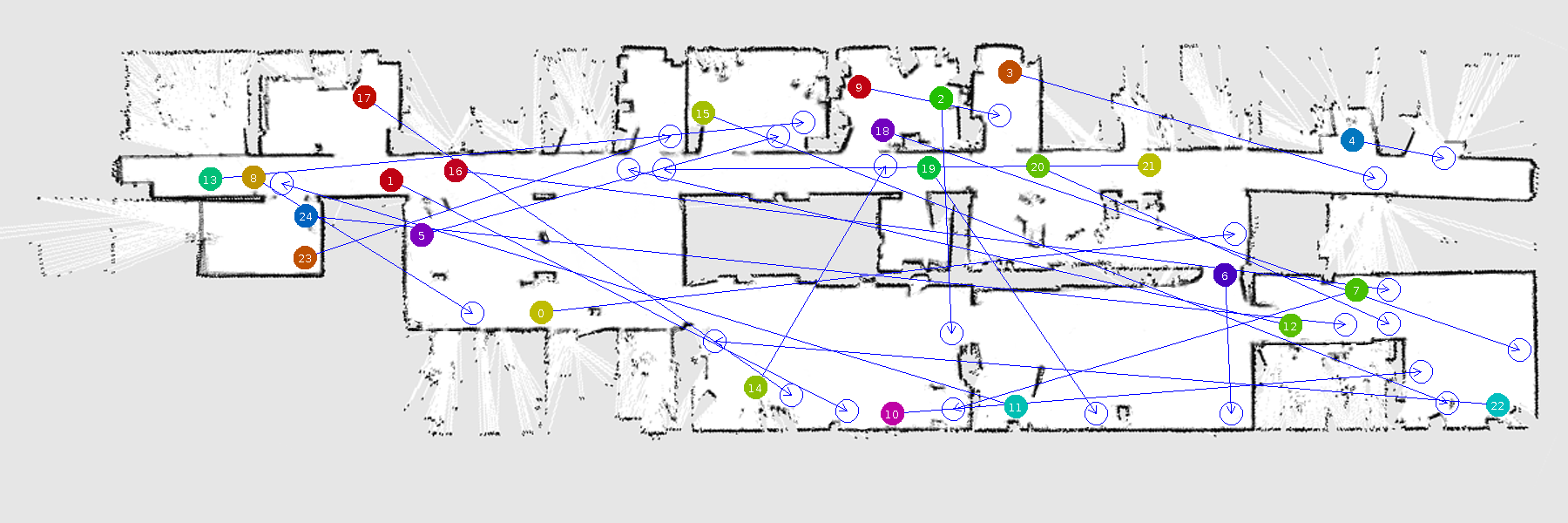}
\par\end{centering}

}
\par\end{centering}

\begin{centering}
\subfloat[\label{fig:Box-roadmap-1-1}Example infrastructure tasks for 25 robots.
Task of each robot shown in blue.]{\begin{centering}
\includegraphics[width=0.9\columnwidth]{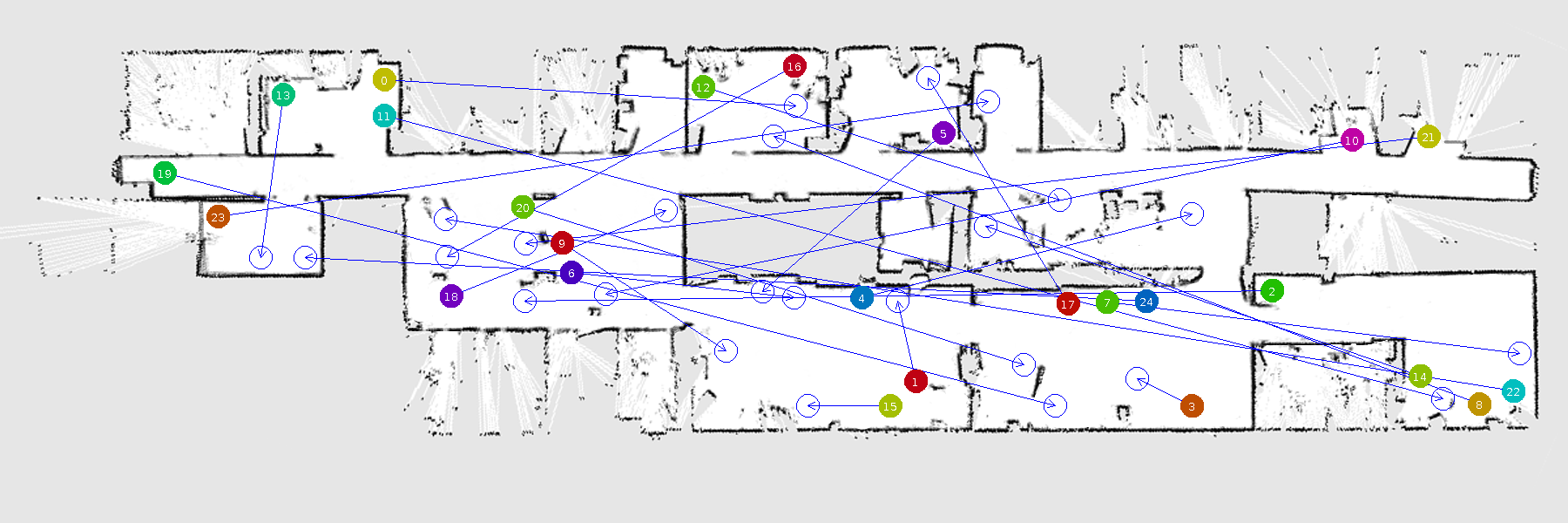}
\par\end{centering}

}
\par\end{centering}

\caption{\label{fig:Corridor-free}Office Corridor environment}
\end{figure}

The \emph{office corridor} environment is based on the laser rangefinder
log of Cartesium building at the University of Bremen. We thank Cyrill
Stachniss for providing the data through the Robotics Data Set Repository~\cite{Radish}.
The environment and the roadmap used for trajectory planning in the
environment are depicted in Figure~\ref{fig:Corridor}. For both
instance sets and each number of robots ranging from $n=1$ to $n=30$,
we generated 25 random problem instances containing the given number
of robots.

\subsubsection*{Warehouse environment}

\begin{figure*}
\begin{centering}
\subfloat[\label{fig:Warehouse-docks} \emph{Warehouse} Environment. The roadmap
used for planning is depicted in gray. Infrastructure endpoints are
shown in red.]{\centering{}\includegraphics[width=0.26\paperwidth]{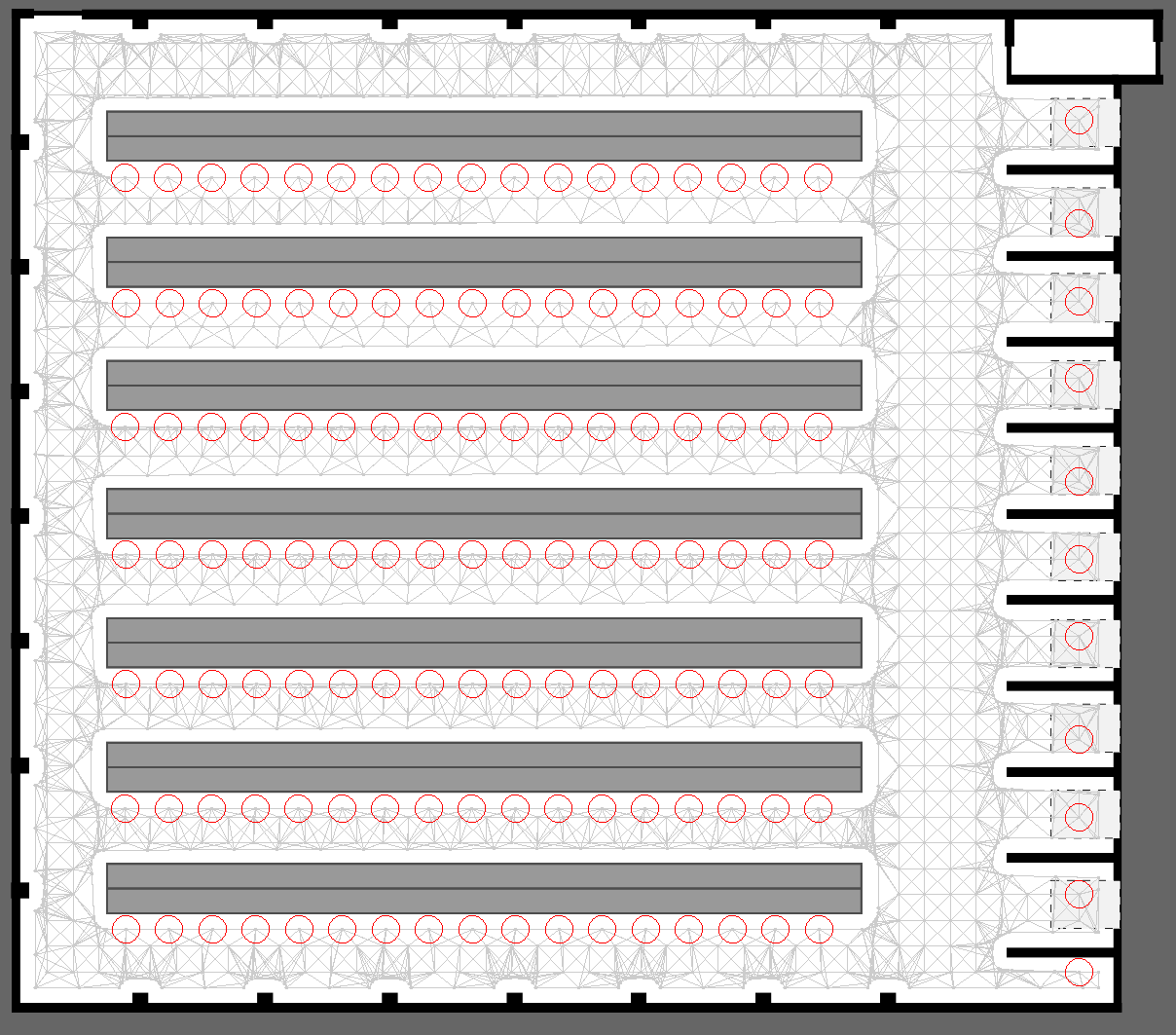}}~\subfloat[\label{fig:Warehouse-free}Example free-formed tasks for 30 robots.
Task of each robot shown in blue.]{\begin{centering}
\includegraphics[width=0.26\paperwidth]{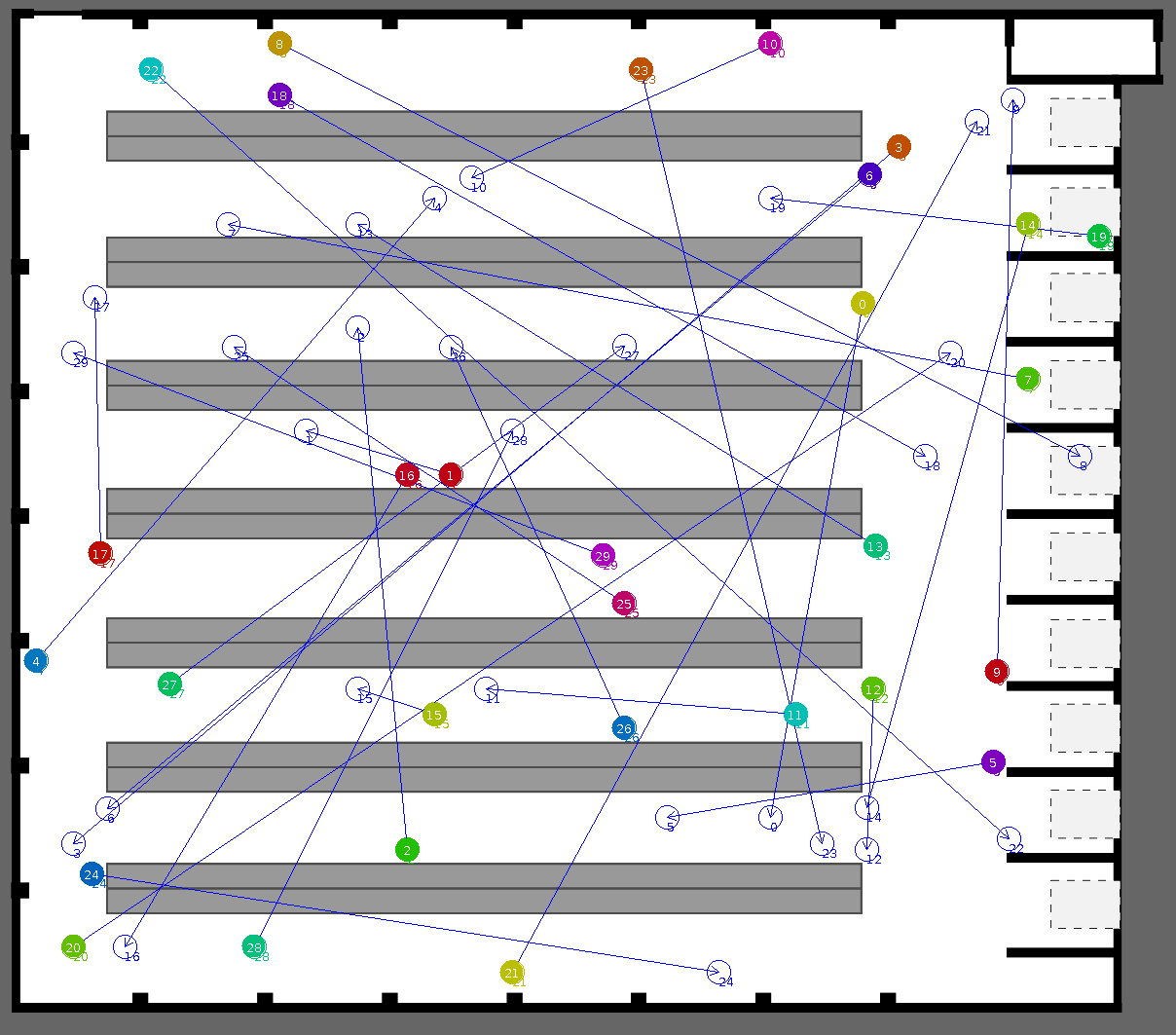}
\par\end{centering}

}~\subfloat[\label{fig:Warehouse-inf}Example infrastructure tasks for 30 robots.
Task of each robot shown in blue.]{\begin{centering}
\includegraphics[width=0.26\paperwidth]{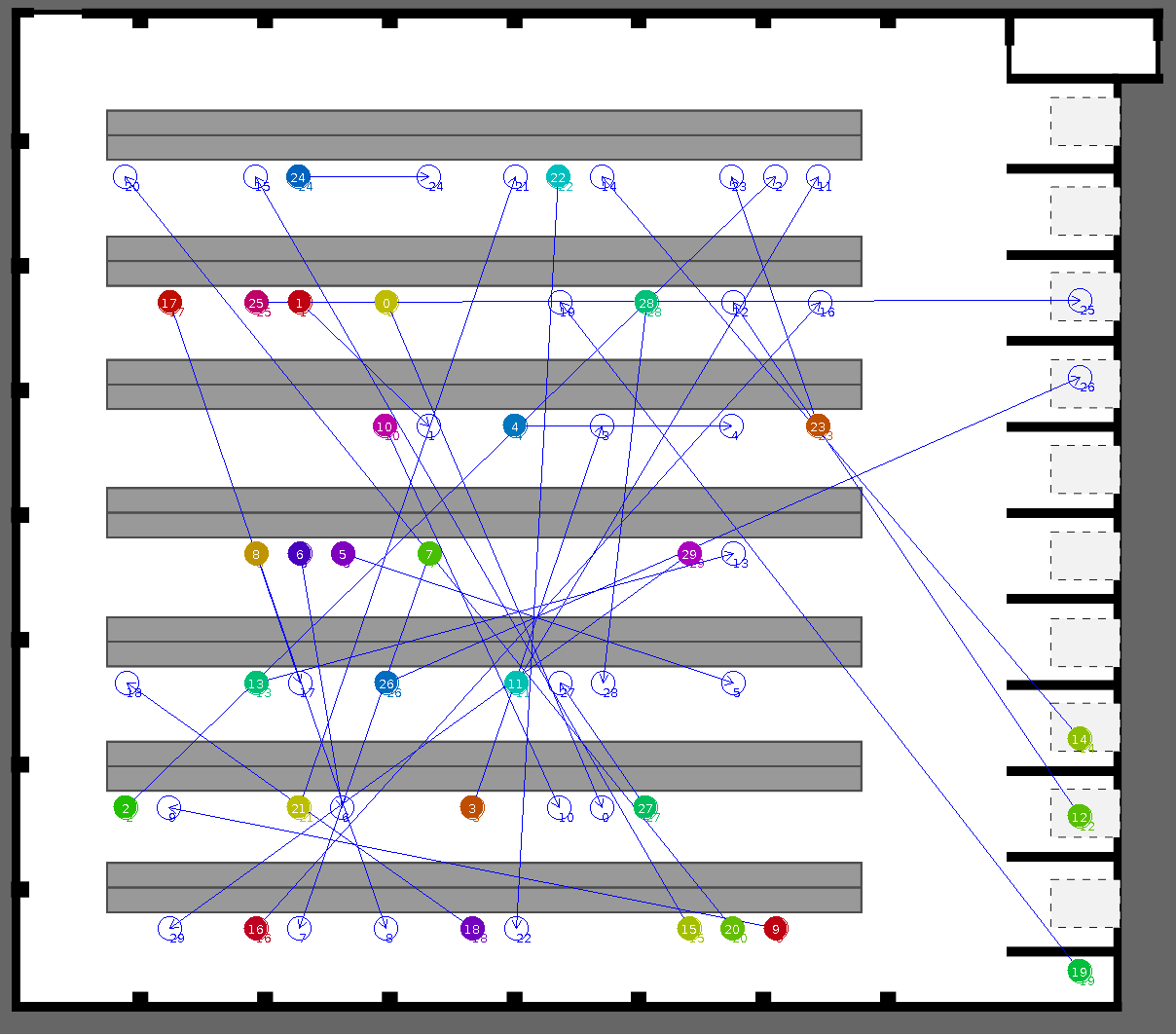}
\par\end{centering}

}
\par\end{centering}

\caption{\label{fig:Warehouse}Warehouse environment}
\end{figure*}

The map of the \emph{warehouse} environment and the roadmap used for
trajectory planning are depicted in Figure~\ref{fig:Warehouse-docks}.
For both instance sets and each number of robots ranging from $n=1$
to $n=60$, we generated 25 random problem instances containing the
given number of robots. The infrastructure tasks instance set represents
a scenario of an automated logistic center where robots move goods
between the gates and the storage shelves.

\subsection*{Results}

\begin{figure*}
\begin{centering}
\subfloat[\textbf{\large{\label{fig:Empty-hall-free-results}Empty hall environment}}]{\begin{centering}
\includegraphics[width=0.27\paperwidth]{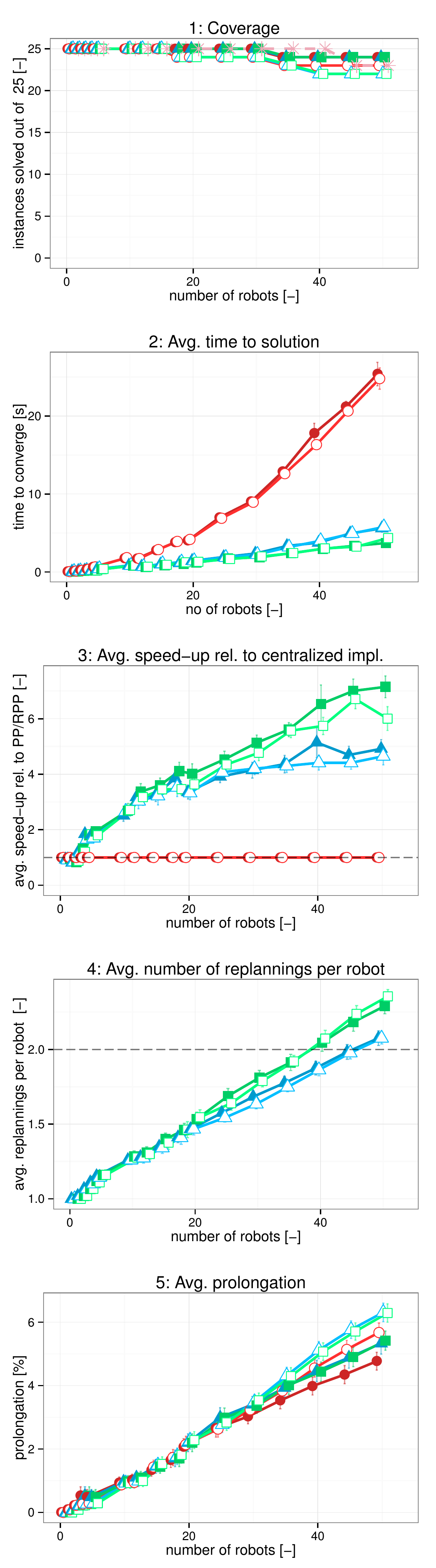}
\par\end{centering}

}\subfloat[\textbf{\large{\label{fig:Corridor-free-results}Office corridor environment}}]{\begin{centering}
\includegraphics[width=0.27\paperwidth]{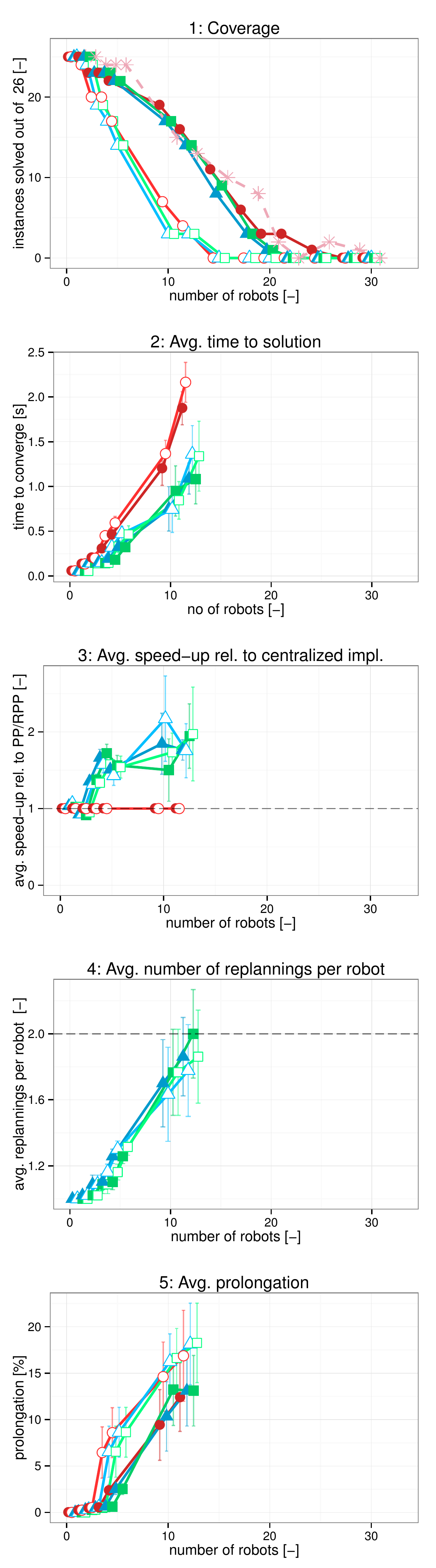}
\par\end{centering}

}\subfloat[\textbf{\large{\label{fig:Warehouse-free-results}Warehouse environment}}]{\begin{centering}
\includegraphics[width=0.27\paperwidth]{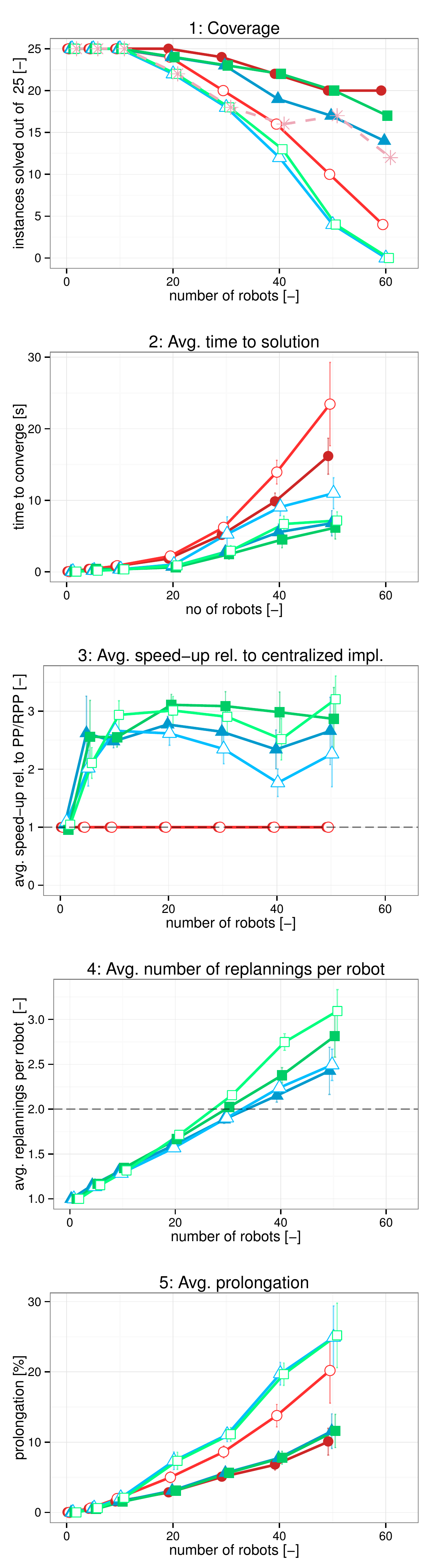}
\par\end{centering}

}
\par\end{centering}

\begin{centering}
\includegraphics[width=0.5\paperwidth]{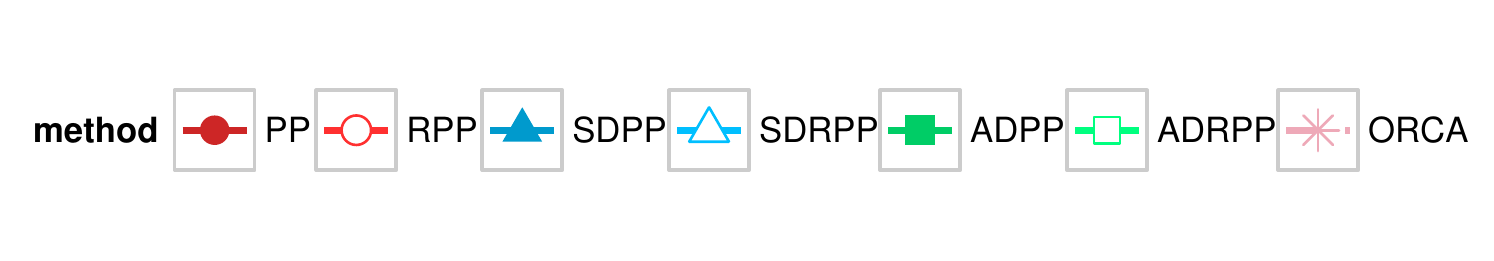}
\par\end{centering}

\centering{}\caption{\textbf{\large{\label{fig:Results-free}}}{\Large{Results: Free-formed
tasks}} (Bars indicate standard error)}
\end{figure*}

\begin{figure*}
\begin{centering}
\subfloat[\textbf{\large{\label{fig:Empty-hall-inf-results}Empty hall environment}}]{\begin{centering}
\includegraphics[width=0.27\paperwidth]{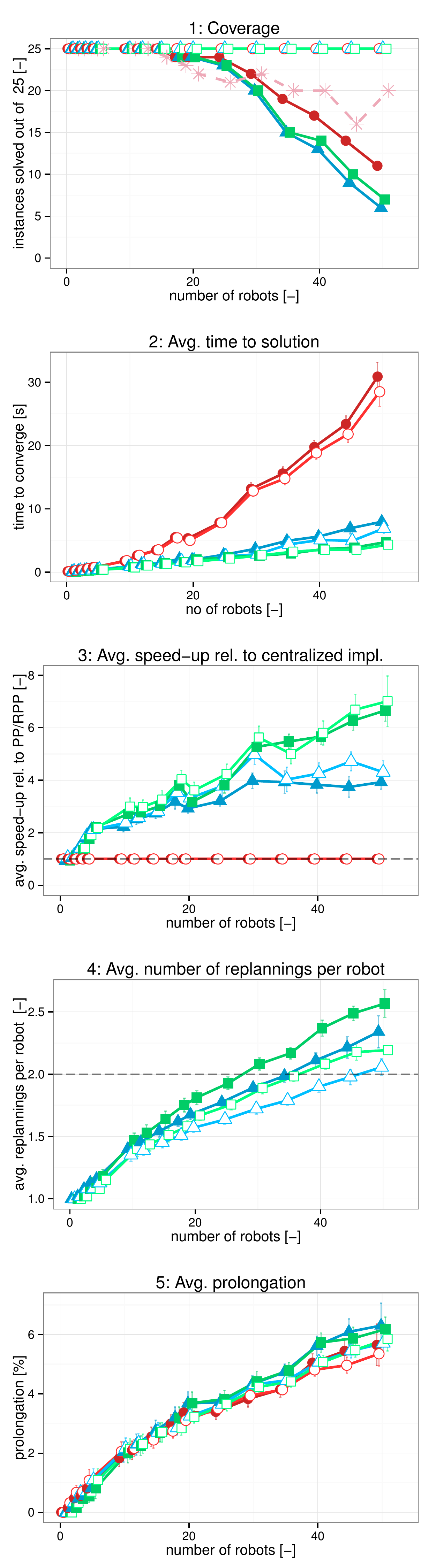}
\par\end{centering}

}\subfloat[\textbf{\large{\label{fig:Corridor-inf-results}Office corridor environment}}]{\begin{centering}
\includegraphics[width=0.27\paperwidth]{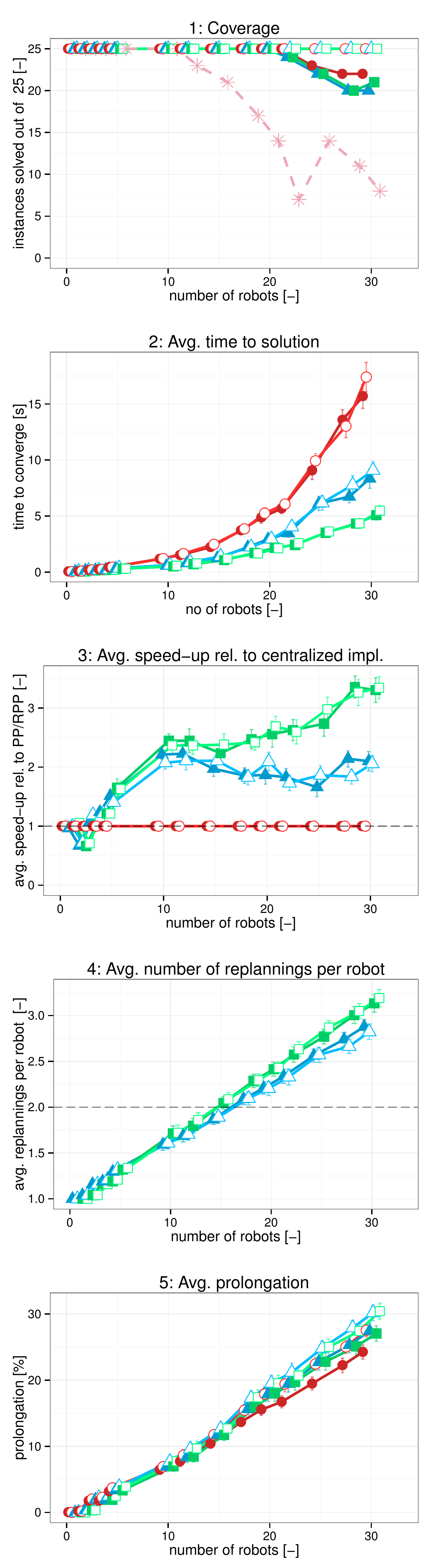}
\par\end{centering}

}\subfloat[\textbf{\large{\label{fig:Warehouse-inf-results}Warehouse environment}}]{\begin{centering}
\includegraphics[width=0.27\paperwidth]{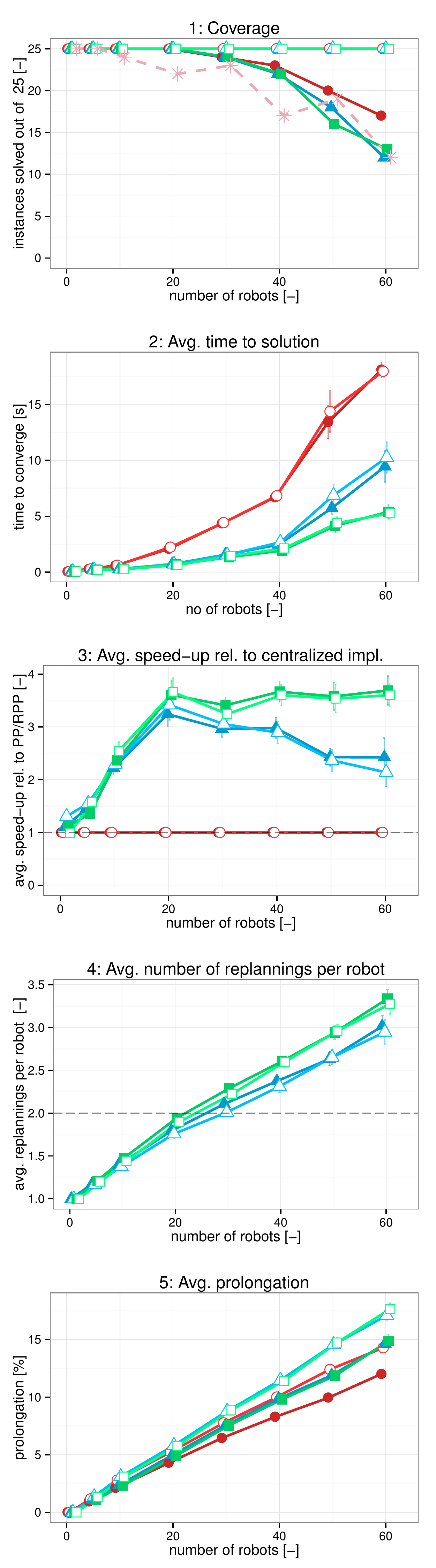}
\par\end{centering}

}
\par\end{centering}

\begin{centering}
\includegraphics[width=0.5\paperwidth]{plots/legend}
\par\end{centering}

\centering{}\caption{\textbf{\large{\label{fig:Results-infrastructure}}}{\Large{Results:
Infrastructure tasks}} (Bars indicate standard error)}
\end{figure*}

The results of the comparison in the three test environments for free-formed
and infrastructure tasks are plotted in Figure~\ref{fig:Results-free}
and Figure~\ref{fig:Results-infrastructure} respectively. 

Note that in the corridor environment and warehouse environment with
free-formed tasks, all tested algorithms exhibit low success-rate
on instances with higher number of robots. Since the plots from these
two instance sets are based only on the few instances that all the
tested algorithms were able to solve, they are shown only for the
sake of completeness and will not be used to draw statistically significant
conclusions.

\subsubsection*{Coverage}

For free-formed tasks, all tested algorithms exhibit incomplete coverage
of the instance space. Generally, RPP-based algorithms solved fewer
instances than the PP-based algorithms. This phenomena can be explained
as follows: The failure of PP-based algorithms is due to Type A (robot
cannot avoid being run-over by a higher-priority robot) or Type B
(all paths to the goal of a robot are blocked by higher-priority robots
at their goal position) conflicts. The failure of a RPP-based algorithms
can be caused either by non-existence of path that avoids start-positions
of higher-priority robots or by Type-B conflict. It is more likely,
however, that a start-avoiding path will not exists (since it is a
static condition that holds at all times) than that a Type B conflict
will occur, thus the lower success rate of RPP-based algorithms.

For infrastructure tasks, the RPP-based instances show full-instance
coverage in accordance with our theoretical analysis. Further, we
can see (best in Figure~\ref{fig:Empty-hall-inf-results}-1) that
some of those instances remain unsolved both by PP-based algorithms
and ORCA.

\subsubsection*{Time to solution/speed-up}

The asynchronous decentralized implementation of both PP and RPP consistently
achieves higher speed-up than the synchronized implementation in accordance
with our prediction. The higher speed-up is exhibited on instances
with higher number of robots, where it is more likely that several
independent conflict clusters will occur. On such instances, it is
often beneficial that the conflict clusters can develop at different
pace and thus converge faster. The phenomena can be seen clearly in
Figures~\ref{fig:Empty-hall-free-results}-3,~\ref{fig:Empty-hall-inf-results}-3,
\ref{fig:Corridor-inf-results}-3, and \ref{fig:Empty-hall-inf-results}-3.

\subsubsection*{Replannings/Communication}

AD-(R)PP broadcast higher number of messages than SD-(R)PP. To see
how this can be explained, suppose that at some point of the computation
new conflicts arise between the trajectory of one particular robot
and the trajectories of two other higher-priority robots. If the two
conflicts occur in a single round, SD-(R)PP solves both conflicts
during one replanning at the end of the round and therefore broadcasts
only a single INFORM message. However, in such a situation AD-(R)PP
may need to replan twice because it triggers replanning immediately
after each of the conflicts is detected and thus it will broadcast
two INFORM messages.

\subsubsection*{Prolongation}

There are two phenomena influencing the quality of returned solutions.
First, RPP-based algorithms generate slightly longer trajectories
than PP-based algorithms. This is due to the fact that RPP preemptively
avoids start positions of the lower-priority robots. Second, decentralized
approaches generate slightly longer trajectories than the centralized
approaches. The reason is the replanning condition used by the decentralized
algorithms. The condition states that a robot should replan its trajectory
only if the trajectory is inconsistent with the trajectories of other
robots. Thus, the robot may receive an updated trajectory from a higher-priority
robot that allows an improvement in its current trajectory, but since
its current trajectory may be still consistent, the robot will not
exploit such an opportunity for an improvement.

\section{\label{sec:Deployment}Deployment}

To verify the applicability of the proposed algorithm in realistic
communication conditions, we deployed the algorithm as a conflict
resolution mechanism in a multi-UAV system. The multi-UAV system we
used as a testbed is being developed as a part of a long-running research
effort in our center~\cite{selecky_2013_aamas_demo}, which is concerned
with development of high-level control algorithms for teams of cooperating
UAVs to autonomously carry out tasks such as patrolling, target tracking
or area surveillance in tactical missions.

\subsection*{Multi-UAV Robotic Testbed\label{sub:Hardware-properties}}

\begin{figure*}
\subfloat[\label{fig:Procerus-fixed-wing-UAV}Top: Unicorn fixed-wing UAV from
Lockheed Martin. Bottom: One of UAVs during the field experiment.]{\begin{centering}
\begin{tabular}{c}
\includegraphics[width=0.25\paperwidth]{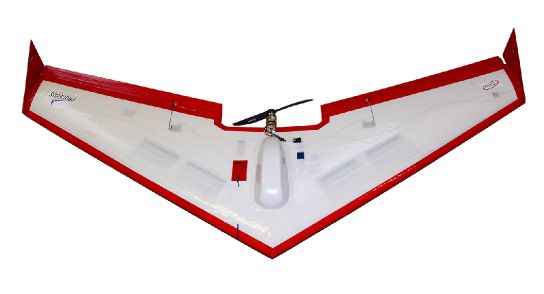}\tabularnewline
\includegraphics[width=0.25\paperwidth]{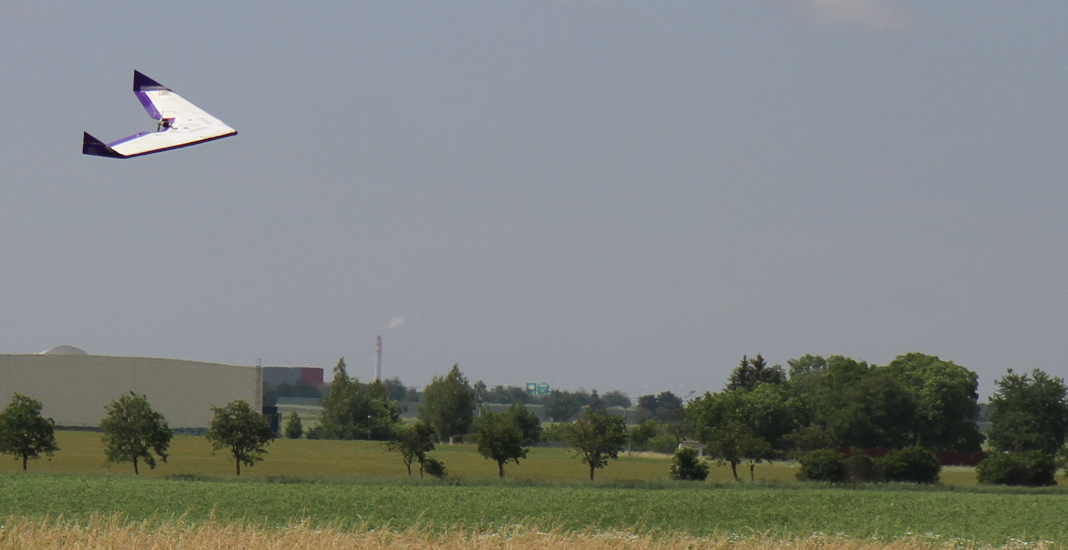}\tabularnewline
\end{tabular}
\par\end{centering}

}\subfloat[Superconflict scenario -- UAV missions \label{fig:Cross-conflict-of-four-mission}]{\begin{centering}
\begin{tabular}{c}
\includegraphics[width=0.25\paperwidth]{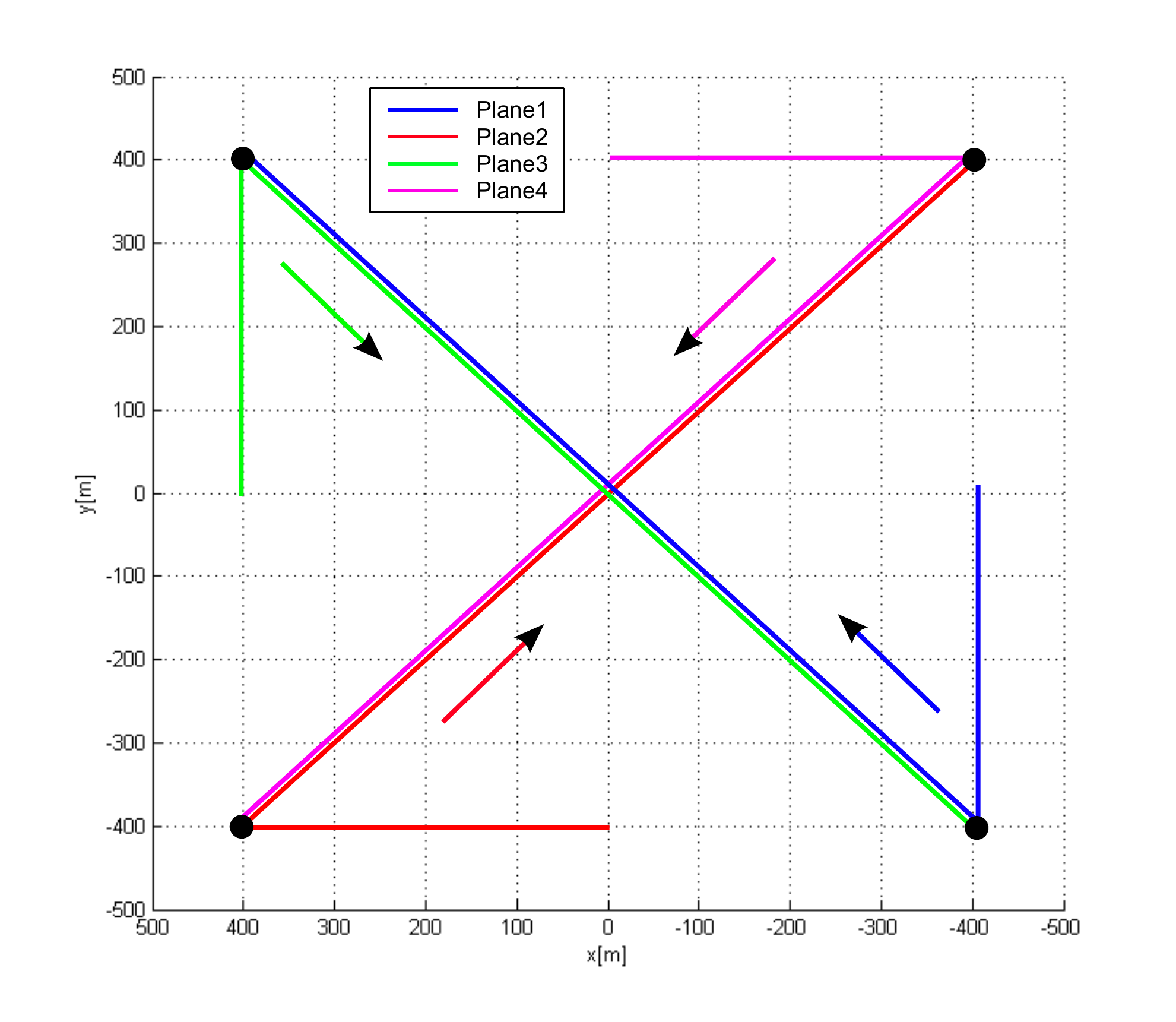}\tabularnewline
\end{tabular}
\par\end{centering}

}\subfloat[Superconflict scenario -- traces of the UAVs \label{fig:Cross-conflict-of-four}]{\begin{centering}
\begin{tabular}{c}
\includegraphics[width=0.25\paperwidth]{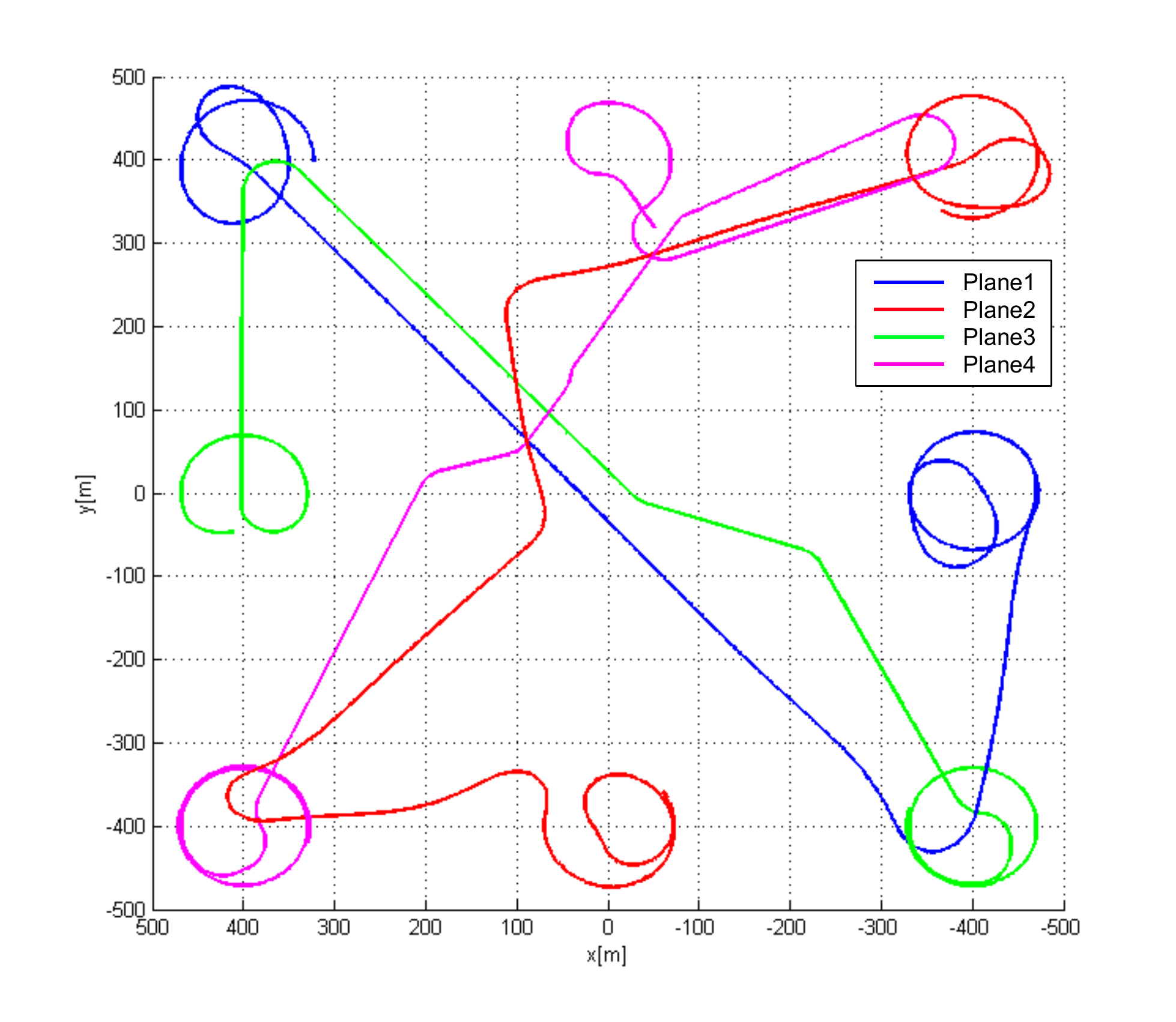}\tabularnewline
\includegraphics[width=0.25\paperwidth]{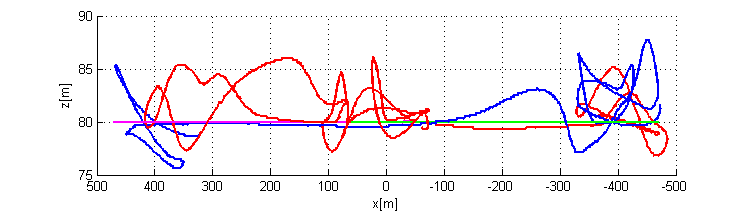}\tabularnewline
\end{tabular}
\par\end{centering}

}

\caption{Deployment to a Multi-UAV system}

\end{figure*}

Our testbed consists of two hardware UAVs and an arbitrary number
of simulated UAVs. The hardware UAVs are based on the Unicorn airframe
equipped with the Kestrel Autopilot from Lockheed Martin (see Figure~\ref{fig:Procerus-fixed-wing-UAV}).
Further, the airborne system is equipped with the Gumstix Overo EarthStorm
embedded computer for on-board computation and Xbee 2.4GHz RF module
to enable direct UAV-to-UAV messaging. Our algorithms are primarily
targeted to large UAV teams. Due to the limited number of physical
UAVs we have at our disposal, a mixed-reality approach~\cite{MRMAS}
is used to scale up.

To address the flying capabilities of the employed UAVs, especially
a relatively high inaccuracy in the plan execution due to unstable
wind, we model the UAV (for the purposes of the trajectory planning
and conflict resolution) as a large cylindric zone around each UAV.
Specifically, we set the radius of the cylinder to 100~m and the
half-height of the cylinder to 10~m. The parameters were chosen empirically
to reflect the autopilot's trajectory tracking precision in difficult
wind conditions. 

An individual UAVs can communicate using a radio link having the shared
capacity 5 kBps in ideal conditions. To keep the latency low and to
use the bandwidth efficiently, the data are transferred in a raw form
and thus the data delivery is not guaranteed.

\subsection*{Closed-Loop AD-PP\label{sub:Closed-Loop-ADPP}}

Four main issues have to be addressed to apply AD-PP as a conflict
resolution algorithm in a real-world multi-UAV system. Firstly, the
whole team or individual UAVs may be re-tasked by an operator and
thus the goals of any of the UAVs may change at any time during the
mission execution. Secondly, the trajectories generated by the ADPP
are often executed imprecisely by the UAVs, especially under unstable
wind conditions. Thirdly, the radio communication channel used in
the system does not guarantee message delivery and messages loss may
occur. Fourthly, during the mission execution, some UAVs may be abruptly
removed from the robotic team, while some new ones can be added. To
address these requirements, we adapted the original AD-PP to work
in a closed-loop fashion. 

In this specific application, a decentralized implementation of classical
prioritized planning turned out to be a better fit than a decentralized
implementation of revised prioritized planning. First, since UAVs
can replan at any time from any point in space, we are not able to
guarantee that $S^{>i}$-avoiding and $G^{<i}$-avoiding satisfying
path will always exists, thus neither of the algorithms can guarantee
completeness. Second, in situations similar to the one depicted in
Figure~\ref{fig:RPP-returns-worse-solution-than-PP}, which are not
uncommon in our case, PP returns shorter solution than RPP. Therefore,
we chose to use PP over RPP as an underlying planning scheme.

Further, we chose asynchronous decentralized implementation of PP
over the synchronized implementation, because in SD-PP the robots
would need to run some form of distributed termination detection algorithm
(e.g.~\cite{mattern1987_distr_termination_detection}) at the end
of each round in order to detect that every other robot has finished
computing in the given round. With unreliable communication channels,
such synchronization is in general impossible to achieve, as was shown
in the famous Two General's problem~\cite{AkkoyunluEH75_TwoGeneralsProblem}.

The pseudo code of the closed-loop version of the AD-PP (called CLAD-PP)
is given in Algorithm~\ref{alg:CLRPP}. In CLAD-PP, an execution
of the trajectories is continuously monitored and a replanning is
invoked if necessary. The replanning is triggered: a) if the robot
is assigned a new task (e.g. by an operator); or b) if the robot has
diverted from its planned trajectory. 

Observe that each such forced replanning triggers a new trajectory
coordination query for the lower-priority robots and thus it may cause
a cascade of replannings for lower-priority UAVs. However, just as
in the standard AD-PP, each of the robots will eventually adapt a
conflict-free trajectory or reports a failure to find one. Because
of the possible message loss, the UAVs broadcast their planned trajectory
not only when it changes, but also periodically onwards. The possible
dynamic team reconfiguration does not allow to wait for a global termination
of the ADPP run, and therefore, a robot starts executing its collision-free
trajectory immediately when it is planned (lines \ref{alg:CLPP_follow1},
\ref{alg:CLPP_follow2}, and \ref{alg:CLPP_follow3}). For the trajectory
generation (\noun{BestTraj} routine) we use an any-time RRT{*}-based~\cite{RRTStar:Karaman.Frazzoli:IJRR11}
spatio-temporal trajectory planner, which is restricted to provide
a solution within one second. 

From a practical point of view, it should be noted that such an extension
was possible due to the asynchronous nature of the AD-PP. Implementing
such a dynamic mechanism with a centralized or a synchronized algorithm
would be much less natural.

\SetKwFunction{clpp}{CLAD-PP}
\SetKwFunction{when}{When}{}
\SetKwProg{periodically}{Periodically}{}{} 
\SetKwProg{when}{When}{}{}
\SetKw{follow}{follow}
\newcommand\mycommfont[1]{\footnotesize\ttfamily\textcolor{gray}{#1}} \SetCommentSty{mycommfont}
\begin{algorithm}
\alg{\clpp}{

	$\pi_{i}\leftarrow\emptyset$\;

	$\ts_{i}\leftarrow\emptyset$\;

	$\pi_{i}\leftarrow$\assertconsistency{$\pi_{i},\mathcal{W},\dobst(\ts_{i})$}\;

	\follow$\pi_{i}$\; \label{alg:CLPP_follow1}

}

\handlemessage{$\mathrm{INFORM}(j,\delta_{j})$}{

	\If{$j<i$}{

	$\ts_{i}\leftarrow\left(\ts_{i}\setminus\left\{ (j,\delta'_{j})\,:\,(j,\delta'_{j})\in\ts_{i}\right\} \right)\cup\left\{ (j,\delta_{j})\right\} $\;

	}

\tcp{plans from the robot's current position}

	$\pi_{i}\leftarrow$ \assertconsistency{$\pi_{i},\mathcal{W},\dobst(\ts_{i})$}\;

	\If{$\pi_{i}=\emptyset$}{

		 report failure and terminate\;

	}

	\follow$\pi_{i}$\;\label{alg:CLPP_follow2}

}

\when{task changes \textbf{or} robot diverted from $\pi_{i}$}{

\tcp{plans from the robot's current position}

	$\pi_{i}\leftarrow$\assertconsistency{$\pi_{i},\mathcal{W},\dobst(\ts_{i})$}\;

	\follow$\pi_{i}$\;\label{alg:CLPP_follow3}

}

\periodically{}{

		\broadcast$\mathrm{INFORM}(i,R_{i}^{\Delta}(\pi))$\;

}

\caption{\label{alg:CLRPP}Asynchronous Decentralized Implementation of Prioritized
Planning, Closed-Loop version}
\end{algorithm}

\subsection*{Superconflict experiment\label{sub:Superconflict-experiment}}

The behavior of the proposed technique within a complex mission is
demonstrated in a so-called ``superconflict'' scenario. We considered
four UAVs with starting positions placed at the corners of a square
and their goals being at the respective diagonal opposite corners.
Hence, all the airplanes are initially in a conflict in the middle
of the square, see Figure~\ref{fig:Cross-conflict-of-four-mission}.
In order to show the behaviour of the CLAD-PP technique more clearly,
the \noun{BestTraj }trajectory planners of the individual UAVs were
constrained to use only a fixed flight altitude and a fixed flight
velocity. 

In this experimental setup, two real UAVs (Plane1 and Plane2) are
attached to a hardware-in-the-loop simulator and two others (Plane3
and Plane4) are simulated. The control algorithms are deployed and
run on the Gumstix on-board computers. The hardware UAVs use the safe
zone radius 110\,m, while the simulated ones use 70~m. The virtual
UAVs are controlled by the identical software as the hardware UAVs;
however, they run in independent virtual machines on a desktop computer.
Both real and simulated UAVs communicate via their XBee radio modules.
The Kestrel autopilot of the hardware UAVs is connected to a high-fidelity
flight simulator Aviones%
\footnote{http://aviones.sourceforge.net/%
} in the hardware-in-the-loop mode. When the mission is started the
UAVs execute the CLAD-PP algorithm to coordinate their motions. The
resulting traces that were recorded during the experiment are shown
in Figure~\ref{fig:Cross-conflict-of-four} and can also be seen
in the attached video. One can see the typical phenomena of prioritized
planning -- the highest-priority Plane 1 keeps its first straight-lane
trajectory, while all other UAVs need to change their first trajectory
to adapt.

\section{Conclusion}

Prioritized planning is a practical approach for multi-robot trajectory
planning. In this paper, we have summarized properties and compared
performance of six different algorithms employing the idea of prioritized
planning. While PP and SD-PP are existing algorithms that have previously
appeared in the literature, the remaining four algorithms are our
novel contributions. 

In particular, we have proposed a revised version of prioritize planning
(RPP) and proved that this algorithm is guaranteed to provide solution
if there is a path for every robot that reaches its goal position,
avoids start positions of lower-priority robots and avoids goal positions
of higher-priority robots. We have shown that this condition is satisfied
if the individual robots move between two endpoints of a valid infrastructure.
The significance of this result lies in the fact that human-made environments
are usually build as valid infrastructures and thus the RPP algorithm
can be used to efficiently find coordinated trajectories for the robots
operating in such environments. We have experimentally demonstrated
that in valid infrastructures, RPP algorithm solves instances that
would be otherwise unsolvable by state-of-the-art techniques such
as the classical prioritized planning or ORCA.

A decentralized algorithm for the multi-robot trajectory coordination
problem is often more desirable because the trajectory planning may
be performed locally by each robot and several robots may plan in
parallel. We proposed a novel asynchronous decentralized implementation
of (revised) prioritized planning scheme and proved that the algorithm
is guaranteed to terminate. Further, the asynchronous decentralized
implementation of revised prioritized planning is guaranteed to provide
a solution under the same conditions as its centralized counterpart
and thus it can be used to reliably plan coordinated trajectories
in valid infrastructures. Further we have shown that in our test environments
the asynchronous approach converges faster than the previously known
synchronized approach. Further, we have demonstrated that the proposed
asynchronous approach is flexible enough to be used as an on-line
mechanism for conflict resolution in a multi-UAV system. 

In future, we plan to study extensions of the presented decentralized
algorithms to open multi-robot systems with local communication.

\subsection*{Acknowledgements}

This research was supported by the Czech Science Foundation (grant
No. 13-22125S) and by the Grant Agency of the Czech Technical University
in Prague grant SGS14/143/OHK3/2T/13. Access to computing and storage
facilities owned by parties and projects contributing to the National
Grid Infrastructure MetaCentrum, provided under the program \textquotedbl{}Projects
of Large Infrastructure for Research, Development, and Innovations\textquotedbl{}
(LM2010005), is greatly appreciated.

\bibliographystyle{plain}
\bibliography{bib}

\begin{lyxcode}

\end{lyxcode}

\end{document}